\documentclass[10pt,twocolumn,letterpaper]{article}

\usepackage{cvpr}
\usepackage{times}
\usepackage{epsfig}
\usepackage{graphicx}
\usepackage{amsmath}
\usepackage{amssymb}

\usepackage[utf8]{inputenc} 
\usepackage[T1]{fontenc}    
\usepackage[draft]{hyperref}       
\usepackage{url}            
\usepackage{booktabs}       
\usepackage{amsfonts}       
\usepackage{nicefrac}       
\usepackage{microtype}      

\usepackage{bm}
\usepackage{url}
\usepackage{float}
\usepackage{amsmath}
\usepackage{amssymb}
\usepackage{amsthm}
\usepackage{xspace}
\usepackage{color}
\usepackage{graphicx}
\usepackage{grffile}
\usepackage{soul}
\usepackage{etoolbox}
\usepackage{chngpage}
\usepackage{environ}
\usepackage{subcaption}
\usepackage{rotating}
\usepackage{multirow}
\usepackage{algorithm}
\usepackage{algorithmicx}
\usepackage[noend]{algpseudocode}
\usepackage{cleveref}
\usepackage{bigints}
\usepackage{mathtools}
\usepackage{wrapfig}

\def\eqref#1{equation~\ref{#1}}

\def\1{\bm{1}}

\def\rvtheta{{\mathbf{\theta}}}

\DeclareMathAlphabet{\mathsfit}{\encodingdefault}{\sfdefault}{m}{sl}
\SetMathAlphabet{\mathsfit}{bold}{\encodingdefault}{\sfdefault}{bx}{n}

\newcommand{\E}{\mathbb{E}}

\DeclareMathOperator*{\argmax}{arg\,max}

\DeclarePairedDelimiter{\norm}{\lVert}{\rVert}

\newtoggle{shortpaper}

\newcommand{\Mod}[1]{\ (\mathrm{mod}\ #1)}

\cvprfinalcopy 

\def\cvprPaperID{****} 

\ifcvprfinal\fi

\newtheorem{theorem}{Theorem}[section]
\newtheorem{conjecture}{Conjecture}[section]

\newtheorem{definition}{Definition}[section]

\newtheorem{prop}{Proposition}

\begin{document}

\title{Extensions and limitations of randomized smoothing for robustness guarantees}

\author{Jamie Hayes\\
University College London\\
{\tt\small j.hayes@cs.ucl.ac.uk}
}

\maketitle
\thispagestyle{empty}

\begin{abstract}

Randomized smoothing, a method to certify a classifier's decision on an input is invariant under adversarial noise, offers attractive advantages over other certification methods. It operates in a black-box and so certification is not constrained by the size of the classifier's architecture. Here, we extend the work of Li et al. \cite{li2018second}, studying how the choice of divergence between smoothing measures affects the final robustness guarantee, and how the choice of smoothing measure itself can lead to guarantees in differing threat models. To this end, we develop a method to certify robustness against any $\ell_p$ ($p\in\mathbb{N}_{>0}$) minimized adversarial perturbation.  We then demonstrate a negative result, that randomized smoothing suffers from the curse of dimensionality; as $p$ increases, the effective radius around an input one can certify vanishes.

\end{abstract}

\section{Introduction}
\label{sec: Introduction}

Image classification is vulnerable to \emph{adversarial examples}.
Given an image classifier $f:\mathbb{R}^n\rightarrow\mathbb{R}^m$ such that the decision function $F = \argmax_i f_i$ classifies an input, $x$, 
correctly as $F(x)=y$, an adversarial example is an input,
$x+\delta$, such that
$F(x+\delta)\neq y$ where $x$ and $x+\delta$ are 
assigned the same label by an oracle classifier, 
$\mathcal{O}$, which is usually
taken to be the human vision system. To preserve oracle classification, 
it is common to minimize the perturbation, $\delta$, with
respect to an $\ell_p$ norm. Constructing a perturbation such that $\norm{\delta}_p \ll \norm{x}_p$, 
will result in an input such that $\norm{x+\delta}_p\approx \norm{x}_p$. With high likelihood $x$ and $x+\delta$ will be visually similar
and $\mathcal{O}$ will classify both correctly.

The vulnerability to adversarial examples requires a suitable
defense. Many empirical defenses have been proposed and subsequently shown to be broken, implying more theoretically grounded techniques to measure robustness are required~\cite{athalye2018obfuscated, carlini2019ami, carlini2019evaluating, engstrom2018evaluating, uesato2018adversarial}. Recently, methods from verification literature have been used to provide guarantees of an
inputs robustness to adversarial perturbations. These methods seek the minimum or a lower bound on the amount of noise required to cause a misclassification. These verification methods are most often tailored to a single $\ell_p$ norm 
for which the defense guarantees robustness. 
A number of defenses \emph{certify} a neural network
is robust to adversarial examples by propagating upper and lower
input bounds throughout the network or by bounding the Lipschitz value of the network~\cite{boopathy2018cnn, dvijotham2018training, gehr2018ai2, gowal2018effectiveness, liu2019certifying, mirman2018differentiable, tsuzuku2018lipschitz, zhang2018efficient}.

Recently, \emph{randomized smoothing}
has been proposed to certify image classifiers to $\ell_0$, $\ell_1$, and $\ell_2$ perturbations~\cite{cohen2019certified, lecuyer2018certified, lee2019stratified, li2018second}.
By constructing a classifier that outputs a label based on a majority vote under repeated addition of Laplacian or Gaussian 
noise, Lecuyer \etal \cite{lecuyer2018certified} found lower bounds to the amount of noise required for misclassification of an input in the $\ell_1$ or $\ell_2$ norm, respectively.
Following this, Li \etal \cite{li2018second} and Cohen \etal \cite{cohen2019certified} provided improved bounds in the $\ell_2$ norm. 
As explained by Cohen \etal \cite{cohen2019certified}, randomized smoothing has attractive advantages
over other certification methods: it is scalable to large classifiers and
makes no assumption about the architecture. 
In this work, we extend the general framework for randomized smoothing as proposed by Li \etal \cite{li2018second}. Firstly, we study how the choice of divergence between inputs smoothed with noise affects the final certificate, and secondly, we study how the choice of smoothing measure itself can lead to guarantees for differing threat models. Concretely, we show how the choice of smoothing measure allows us to extend randomized smoothing to any $\ell_p$ norm ($p\in\mathbb{N}_{>0}$), showing we can certify inputs with non-vacuous 
bounds over a range of $\ell_p$ norms with small $p$ values. We then show that randomized smoothing fails to certify meaningfully large radii around inputs as $p$ increases.


\section{Certified defenses}
\label{sec: certification_defenses}

In this section, we discuss related work on certified defenses to adversarial examples, introduce extensions to randomized smoothing approaches to certified defenses, and
provide a method to compute a certified robust area around an input
under \emph{any} $\ell_p$ norm attack, where $p\in\mathbb{N}_{>0}$.

\subsection{Background on certified defenses}

The vulnerability of empirical defenses to adversarial examples has driven
the need for formal guarantees of robustness.
We define \emph{certified robustness} as a guarantee that the decision
of a classifier is preserved
within an $\epsilon$-ball around an input, and we refer to size of this $\epsilon$-ball as the \emph{certified radius}. Formal 
methods can be separated into \emph{complete} and
\emph{incomplete} methods. Complete methods such
as Satisfiability Modulo Theory (SMT)~\cite{carlini2018ground, ehlers2017formal, katz2017reluplex} or Mixed-Integer
Programming (MIP)~\cite{bunel2018piecewise, cheng2017maximum, weng2018towards} provide
 exact robustness bounds but are expensive to 
implement. Incomplete methods solve a convex 
relaxation of the verification problem. The
bounds given by incomplete methods can be loose
 but are quicker to find than exact bounds~\cite{boopathy2018cnn, dvijotham2018training, gehr2018ai2, gowal2018effectiveness, liu2019certifying, mirman2018differentiable, zhang2018efficient}. 

Lecuyer \etal \cite{lecuyer2018certified} developed the certification technique, referred to as \emph{randomized smoothing}, by noticing a connection between differential privacy~\cite{dwork2006calibrating} and robustness, and show that robustness can be proven under
concentration measures of classification under noise. This work was expanded upon by Lee \etal \cite{lee2019stratified}, Li \etal \cite{li2018second}, and Cohen \etal \cite{cohen2019certified}, who found improved robustness guarantees in the $\ell_0$, $\ell_1$, and $\ell_2$ norms, respectively. Similarly to this work, Dvijotham \etal \cite{Dvijotham2020A} developed a general framework for randomized smoothing that can handle arbitrary smoothing measures and so find robustness guarantees in any $\ell_p$ norm. In concurrent work, Blum \etal \cite{blum2020random}, Kumar \etal \cite{kumar2020curse}, and Yang \etal \cite{yang2020randomized} also show that randomized smoothing may be unable to find robustness guarantees in the $\ell_{\infty}$ norm. Most related to this work are the findings of  Kumar \etal \cite{kumar2020curse}, who also use a generalized Gaussian distribution for smoothing and show that the certified radius in an $\ell_p$ norm decreases as $\mathcal{O}(\nicefrac{1}{d^{\frac{1}{2}-\frac{1}{p}}})$, where $d$ is the dimensionality of the data.

\subsection{Certification via randomized smoothing}
\label{ssec:randomized_smoothing}

Here, we expand on how robustness guarantees can be found through randomized smoothing.

\begin{table*}[t]
\caption{$\ell_2$ certified radius when using different divergences.}
\vspace{0.3em}
\centering
\label{tab:distances_guarantees}
\resizebox{\textwidth}{!}{%
\begin{tabular}{lccc}
\toprule
\multirow{ 2}{*}{Distance} & $d(Q, P)\geq $  & \multirow{ 2}{*}{$d(\mathcal{N}(x, \sigma^2), \mathcal{N}(x', \sigma^2))$} & Certified radius\\
& (when $\argmax_i q_i \neq \argmax_i p_i$) & & (for $\norm{x - x'}_2<\epsilon$) \\
\midrule \vspace{0.1em} \\
$d_{KL}(Q,P)=\sum_{i=1}^k q_i\log\frac{q_i}{p_i}$ & $-\log(2\sqrt{p_1p_2} + 1 - p_1 - p_2)$ &
$\frac{1}{\sigma^2}\norm{x - x'}_2^2$  & 
$\sqrt{-\sigma^2\log(2\sqrt{p_1p_2} + 1 - p_1 - p_2)}$                       \vspace{1em} \\
$d_{H^2}(Q,P)=\frac{1}{2}\sum_{i=1}^k (\sqrt{q_i} - \sqrt{p_i})^2$ & $1 - \sqrt{1-\frac{(\sqrt{p_1} - \sqrt{p_2})^2}{2}}$ & $1-e^{-\frac{\norm{x - x'}_2^2}{8\sigma^2}}$ & $\sqrt{-8\sigma^2\log(\sqrt{1-\frac{(\sqrt{p_1} - \sqrt{p_2})^2}{2}})}$ \vspace{1em}\\
 $d_{\chi^2}(Q,P)=\sum_{i=1}^k \frac{(q_i - p_i)^2}{p_i}$ & $\frac{(p_1-p_2)^2}{(p_1 + p_2) - (p_1 - p_2)^2}$ & $e^{\frac{\norm{x - x'}_2^2}{\sigma^2}} - 1$ & $\sqrt{\sigma^2\log(\frac{p_1 + p_2}{(p_1 + p_2) - (p_1 - p_2)^2}})$ \vspace{1em} \\
 $d_B(Q,P)=-\log(\sum_{i=1}^k \sqrt{q_i p_i})$ & $-\log\big(\frac{(\sqrt{p_1}+\sqrt{p_2})^2 + 2(1-p_1-p_2)}{\sqrt{2(2\sqrt{p_1p_2}+2-p_1-p_2)}}\big)$
  &  $\frac{1}{8\sigma^2}\norm{x - x'}_2^2$ &  $\sqrt{-8\sigma^2\log\big(\frac{(\sqrt{p_1}+\sqrt{p_2})^2 + 2(1-p_1-p_2)}{\sqrt{2(2\sqrt{p_1p_2}+2-p_1-p_2)}}\big)}$ \vspace{1em} \\ 
 $d_{TV}(Q,P)=\frac{1}{2}\sum_{i=1}^k |q_i - p_i|$ & $\frac{|p_1-p_2|}{2}$ & $2\Phi(\frac{\norm{x - x'}_2}{2\sigma}) - 1$ & $2\sigma\Phi^{-1}(\frac{|p_1-p_2|}{2} + \frac{1}{2})$ \vspace{1em} \\
\bottomrule
\end{tabular}%
}
\end{table*}

\noindent\textbf{Problem statement.} Given an input
$x\in\mathcal{X}$ such that $\argmax_i f_i(x)=y$, find the maximum $\epsilon$
such that $\forall x'\in\mathcal{X}$, $d(x,x') < \epsilon \implies \argmax_i f_i(x')=y$, given a distance function $d:\mathcal{X}\times \mathcal{X} \rightarrow \mathbb{R}^{+}$.

This can be cast as an optimization problem, given by

\begin{align}
\begin{split}
& \max_{x'\in\mathcal{X}} d(x, x') \\ & \text{ subject to } \argmax_i f_i(x')=y
\end{split}
\end{align}

In general, solving the above formulation is difficult, however randomized smoothing, introduced by Lecuyer \etal \cite{lecuyer2018certified}, can be used to solve a relaxed version of this problem. Namely, the aim is to solve

\begin{align}
\begin{split}
& \max_{x'\in\mathcal{X}} d(x+\rvtheta, x' + \theta) \\ & \text{ subject to } \E[\argmax_i f_i(x'+\rvtheta)]=y,
\end{split}
\end{align}

where $\rvtheta$ is a sample from a smoothing measure, $\mu$, and $d$ is now taken to be a suitable divergence or distance measure between random variables.  For example, Li \etal \cite{li2018second} take $\mu$ to be the centered Gaussian, $\mathcal{N}(0, \sigma^2)$. Since Gaussians belong to the location-scale family of distributions, we can treat $x$ and $x'$ as constants and so, $x+\rvtheta$ and $x' +\rvtheta$  can be treated as random variables from distributions $\mathcal{N}(x, \sigma^2)$ and $\mathcal{N}(x', \sigma^2)$, respectively. We can use well known properties of divergences of Gaussians to represent $d(x+\rvtheta, x'+\rvtheta)$ in terms of the $\ell_2$ norm difference of their means. Specifically,
$d(x+\rvtheta, x'+\rvtheta)$ can be represented as a function of $\norm{x-x'}_2$ and $\sigma$, for common divergences such as the Rényi and KL divergences. However, we must still solve the problem of ensuring $\E[\argmax_i f_i(x'+\rvtheta)]=y$. Given a chosen divergence, Li \etal \cite{li2018second} approach this problem by finding a lower bound between two multinomial distributions, $P$ and $Q$, in terms of the two largest probabilities of $P$, when $\argmax_i P_i \neq \argmax_i Q_i$. This shows that any distribution, $Q$, for which $P$ and $Q$ agree on the index of the top probability, the divergence between $P$ and $Q$ must be smaller than this lower bound. We denote this lower bound by $h(p_1, p_2)$, where $p_1, p_2$ represent the top two probabilities from $P$. Given this lower bound Li \etal \cite{li2018second}, solve the following problem

\begin{align}
\begin{split}
&\max_{x'\in\mathcal{X}} d(f(x+\rvtheta), f(x' + \theta)) \\ &\text{ subject to } d(f(x+\rvtheta), f(x' + \theta))\leq h(p_1, p_2)
\end{split}
\end{align}

This can be efficiently solved by finding an upper bound to the Lagrangian relaxed problem

\begin{align}
\begin{split}
& \max_{\lambda\leq0,x'\in\mathcal{X}} d(f(x+\rvtheta), f(x' + \theta)) \\
&\quad\quad\quad+ \lambda (h(p_1, p_2) - d(f(x+\rvtheta), f(x' + \theta)))
\end{split}
\end{align}
\begin{align}
&= \max_{\lambda\leq0,x'\in\mathcal{X}} (1-\lambda)d(f(x+\rvtheta), f(x' + \theta)) + \lambda h(p_1, p_2) \label{eq: lagrange0}\\
&= \max_{\lambda\geq0,x'\in\mathcal{X}} (1+\lambda)d(f(x+\rvtheta), f(x' + \theta)) - \lambda h(p_1, p_2) \label{eq: lagrange1}\\
&\leq \max_{\lambda\geq0,x'\in\mathcal{X}} (1+\lambda)d(x+\rvtheta, x' + \theta) - \lambda h(p_1, p_2) \label{eq: lagrange2}\\
&= \max_{\lambda\geq0,x'\in\mathcal{X}} (1+\lambda)g(\norm{x-x'}_2, \sigma) - \lambda h(p_1, p_2),  \label{eq: lagrange3}
\end{align}

where in \cref{eq: lagrange2}, we use the data processing inequality property of divergences, and in \cref{eq: lagrange3}, we use the fact that for many common divergences, we can represent the divergence between two Gaussians as a function of the $\ell_2$ norm of their means and their standard deviation, which we denote by $g(\norm{x-x'}_2, \sigma)$.

By choosing $d:\mathcal{X}\times \mathcal{X} \rightarrow \mathbb{R}^{+}$ to be the Rényi divergence, we recover the results of Li \etal \cite{li2018second} with 
\begin{align}
    &g(\norm{x-x'}_2, \sigma) = \frac{\alpha\norm{x-x'}_2^2}{2\sigma^2}\\
    &h(p_1, p_2) = -\log\Big(1-p_1-p_2 +  2\big(\frac{1}{2}(p_1^{1-\alpha} + p_2^{1-\alpha})\big)^{\frac{1}{1-\alpha}}\Big) \label{eq: renyi_lb}
\end{align}
 Thus, for any $x'\in \mathcal{X}$ with $\norm{x-x'}_2 < \epsilon$ we can guarantee the classifier, $f$, will not change it's decision for any $\epsilon$ smaller than

\begin{align}
\label{eq:lagrange4}
\begin{split} 
\max_{\lambda\geq0}\Bigg(\sup_{\alpha>1}\bigg(-\frac{\lambda2\sigma^2}{(1+\lambda)\alpha}&\log\Big(1-p_1-p_2 + \\&2\big(\frac{1}{2}(p_1^{1-\alpha} + p_2^{1-\alpha})\big)^{\frac{1}{1-\alpha}}\Big)\bigg)\Bigg)^{\frac{1}{2}}\\
\end{split} 
\end{align}
\vspace{-0.25cm}
\begin{align}
\begin{split}
=\Bigg( \sup_{\alpha>1}\bigg(-\frac{2\sigma^2}{\alpha}&\log\Big(1-p_1-p_2 + \\ &2\big(\frac{1}{2}(p_1^{1-\alpha} + p_2^{1-\alpha})\big)^{\frac{1}{1-\alpha}}\Big)\bigg)\Bigg)^{\frac{1}{2}}
\end{split} 
\end{align}

Clearly, this framework for certifying inputs is general and extends to different choices of divergence. In the next section, we explore divergences beyond Rényi divergence and show this choice affects the certified radius, given a Gaussian smoothing measure.

\subsection{Certification guarantees against $\ell_2$ perturbations for common divergences}

\begin{figure}[t!]
\centering
  \includegraphics[width=1.0\linewidth]{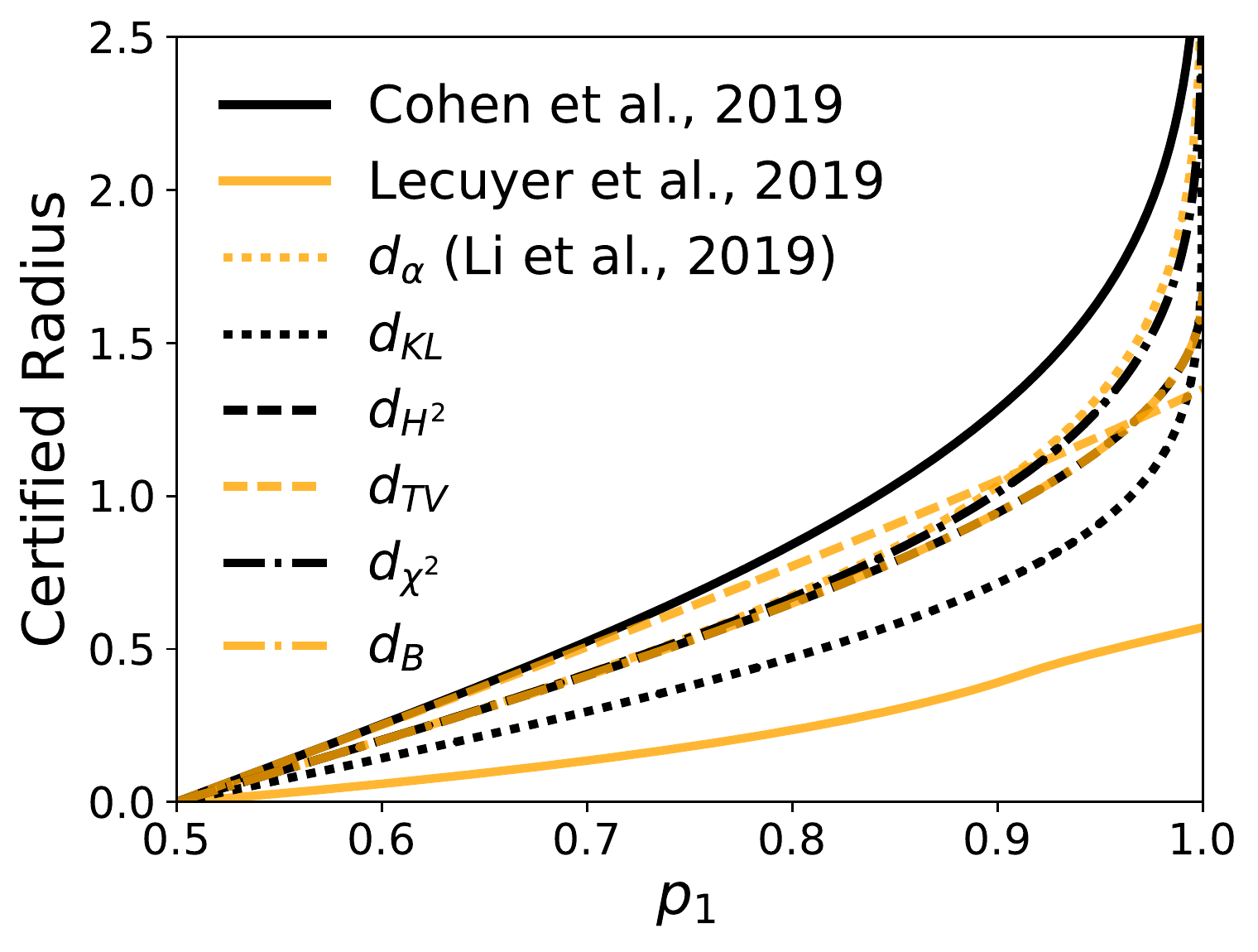}
\caption{Comparison of the certified radius against perturbations targeting the $\ell_2$ norm, for different divergences, as a function of the top predicted probability, $p_1$, with $\sigma=1$.}
\label{fig: l2_divergence_comp}
\end{figure}

Li \etal \cite{li2018second} show that, given two distributions, $P$ and $Q$, with different indexes for the top probability, a lower bound of
the Rényi divergence (denoted by $d_\alpha$)
is given by \cref{eq: renyi_lb}. 
We extend this line of reasoning to find lower bounds for the KL divergence ($d_{KL}$), Hellinger distance ($d_{H^2}$), (Neyman) chi-squared distance ($d_{\chi^2}$), Bhattacharyya distance ($d_{B}$), and total variation distance ($d_{TV}$). Proofs of these lower bounds are given in \cref{app: lb_r_kl}. To find a certified radius of a classifier's decision around an input, we find the distances between Gaussian measures with respect to each of these divergences. These are both represented in \cref{tab:distances_guarantees} along with the certification guarantee in the $\ell_2$ norm. We visualize the trade-off in certified radius around an input in \cref{fig: l2_divergence_comp} for a hypothetical binary classification task as a function of the classifier's top output probability, $p_1$. As well as including the certified radii derived from the aforementioned divergences, we include the certified radii for the $\ell_2$ norm found by Lecuyer \etal \cite{lecuyer2018certified} and Cohen \etal \cite{cohen2019certified} approaches. 
Lecuyer \etal \cite{lecuyer2018certified} find a certified radius against $\ell_2$ perturbations given by $\sup _{0<\beta \leq \min(1, \frac{1}{2} \log \frac{p_1}{p_2})} \frac{\sigma \beta}{\sqrt{2 \log \left(\frac{1.25(1+\exp (\beta))}{p_1-\exp (2 \beta) p_2}\right) }}$, 
while Cohen \etal \cite{cohen2019certified} give a tight robustness guarantee for $\ell_2$ perturbations of the form $\frac{\sigma}{2}\left(\Phi^{-1}\left(p_1\right)-\Phi^{-1}(p_2)\right)$. 

Clearly, all choices of distance metrics dominate the certificates found using the Lecuyer \etal \cite{lecuyer2018certified} method, and for values of $p_1$ close to $\nicefrac{1}{2}$, $d_{TV}$ is approximately equal to the tight Cohen \etal \cite{cohen2019certified} guarantee. However, the certified radius found using $d_{TV}$ is linear with respect to the top predicted probability, and so becomes a weaker guarantee for larger probabilities. Robustness guarantees provided by Rényi and chi-squared divergences are approximately equal; a finer-grained visualization of the difference between these two divergences is given in \cref{app: l2_chi_renyi_compare}.

We formalize the trade-offs between different choices of divergences with the following proposition.

\begin{prop}
\label{prop_main}
Let $\epsilon_{d_{KL}}, \epsilon_{d_{\chi^2}}, \epsilon_{d_{H^2}}, \epsilon_{d_{B}}, \epsilon_{d_{\alpha}}$, and  $\epsilon_{\text{\cite{lecuyer2018certified}}}$, denote the certificates found using $d_{KL}, d_{\chi^2}, d_{H^2}, d_{B}, d_{\alpha}$, and the Lecuyer et al. \cite{lecuyer2018certified} approach, respectively. Then, the following holds 
\begin{enumerate}
    \item $\forall p_1\in(\frac{1}{2}, 1)$, $\epsilon_{d_{\alpha}} > \epsilon_{d_{\chi^2}}$.
    \item $\forall p_1\in(\frac{1}{2}, 1)$, $\epsilon_{d_{\chi^2}} > \epsilon_{d_{KL}}$.
    \item $\forall p_1\in(\frac{1}{2}, 1)$, $\epsilon_{d_{\chi^2}} > \epsilon_{d_{H^2}}$.
    \item $\forall p_1\in[\frac{1}{2}, 1]$, $\epsilon_{d_{B}} = \epsilon_{d_{H^2}}$.
    \item $\forall p_1\in(\frac{1}{2}, 0.998)$, $\epsilon_{d_{H^2}} > \epsilon_{d_{KL}}$.
    \item $\forall p_1\in(\frac{1}{2}, 1)$, $\epsilon_{d_{KL}} > \epsilon_{\text{\cite{lecuyer2018certified}}}$.
\end{enumerate}
\end{prop}

\begin{proof}
See \cref{app: prop_proof}.
\end{proof}

\Cref{prop_main} defines a strict hierarchy, and so informs us of the best divergence one can use to certify an input against $\ell_2$ perturbations using the Li \etal \cite{li2018second} approach.

\begin{figure*}[t!]
\centering
\begin{subfigure}{.248\textwidth}
  \centering
  \includegraphics[width=\linewidth]{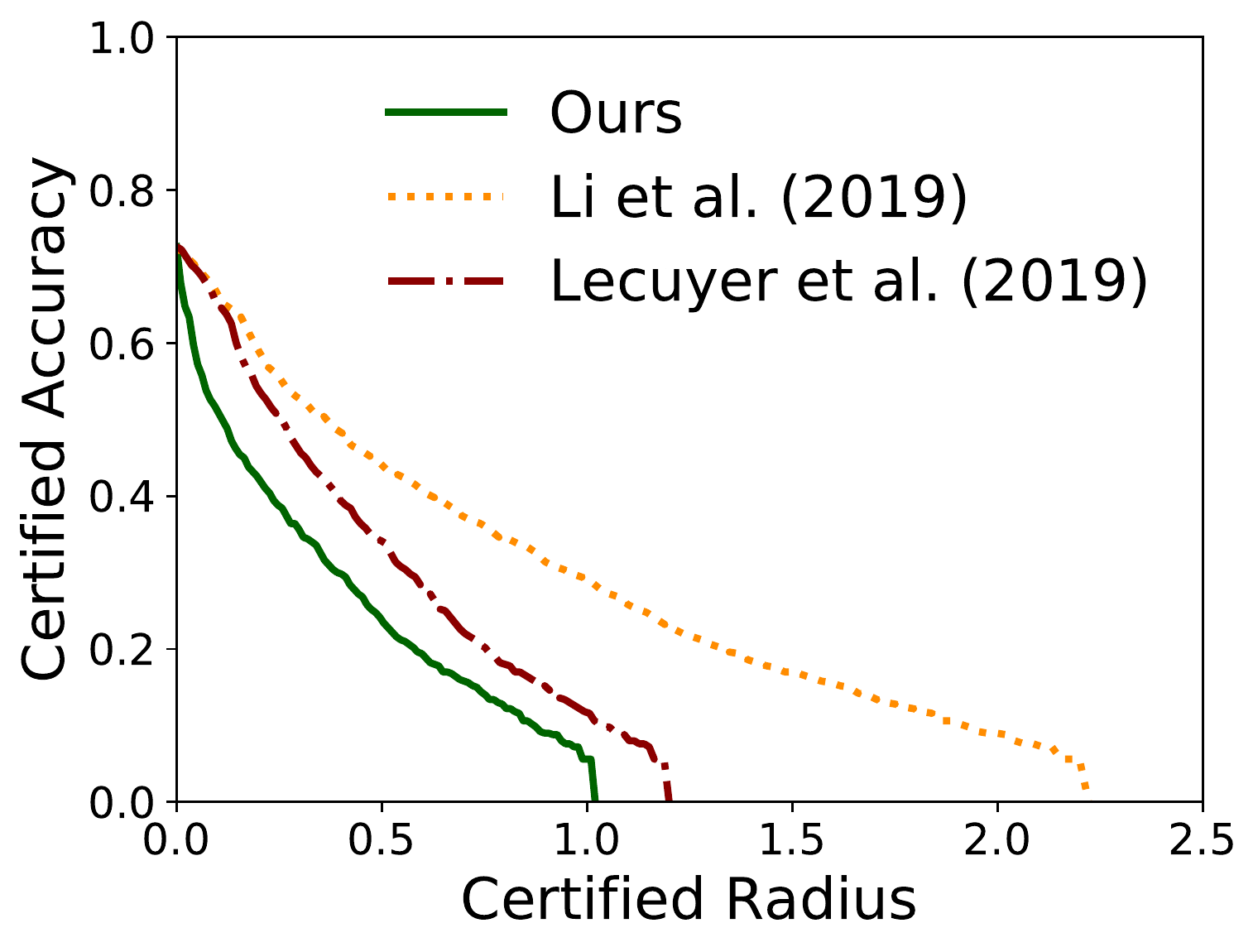}
  \caption{CIFAR-10, $\ell_1$}
  \label{fig:l1_cert_cifar10}
\end{subfigure}%
\begin{subfigure}{.248\textwidth}
  \centering
  \includegraphics[width=\linewidth]{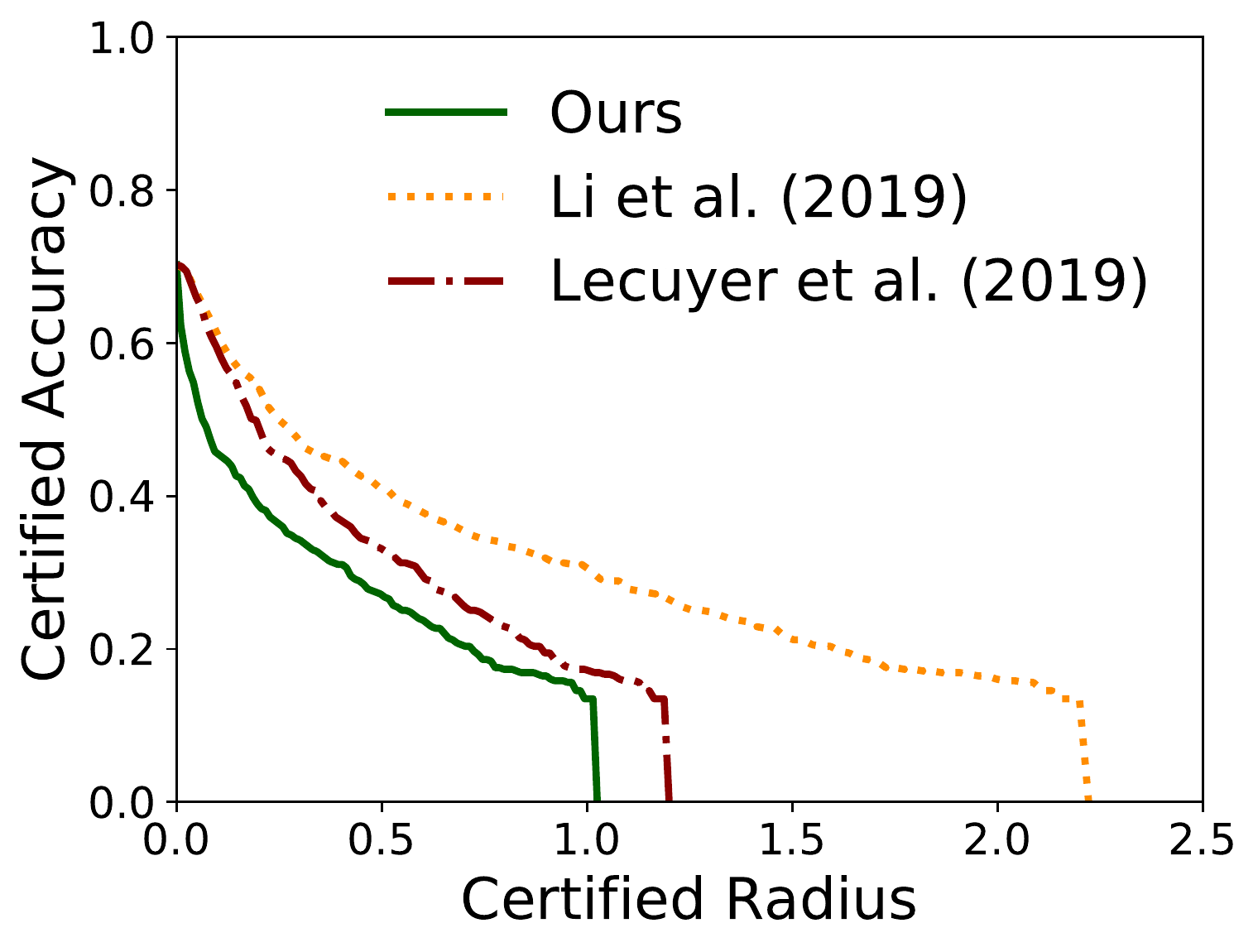}
  \caption{ImageNet, $\ell_1$}
  \label{fig:l1_cert_imagenet}
\end{subfigure}
\begin{subfigure}{.248\textwidth}
  \centering
  \includegraphics[width=\linewidth]{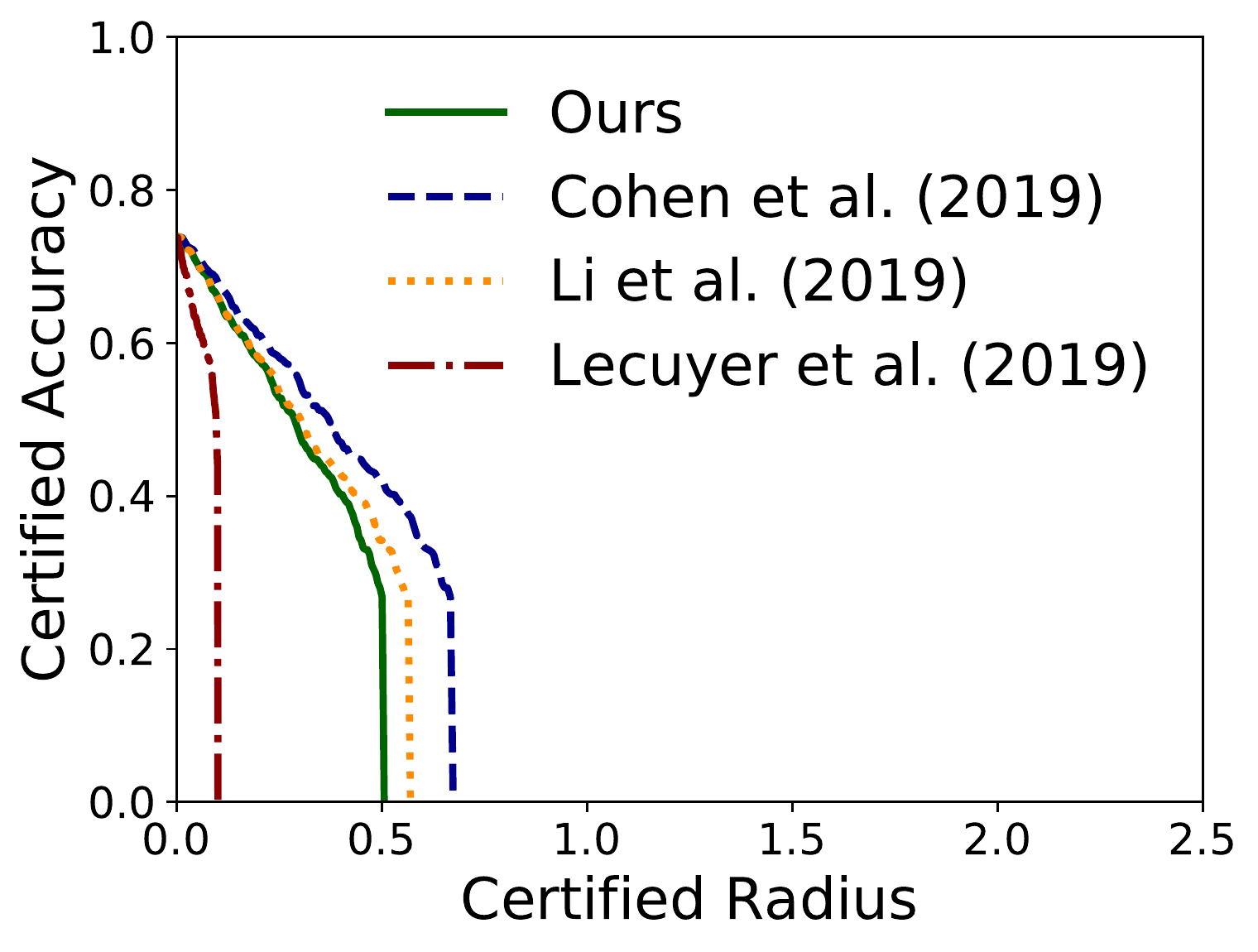}
  \caption{CIFAR-10, $\ell_2$}
  \label{fig:l2_cert_cifar10}
\end{subfigure}%
\begin{subfigure}{.248\textwidth}
  \centering
  \includegraphics[width=\linewidth]{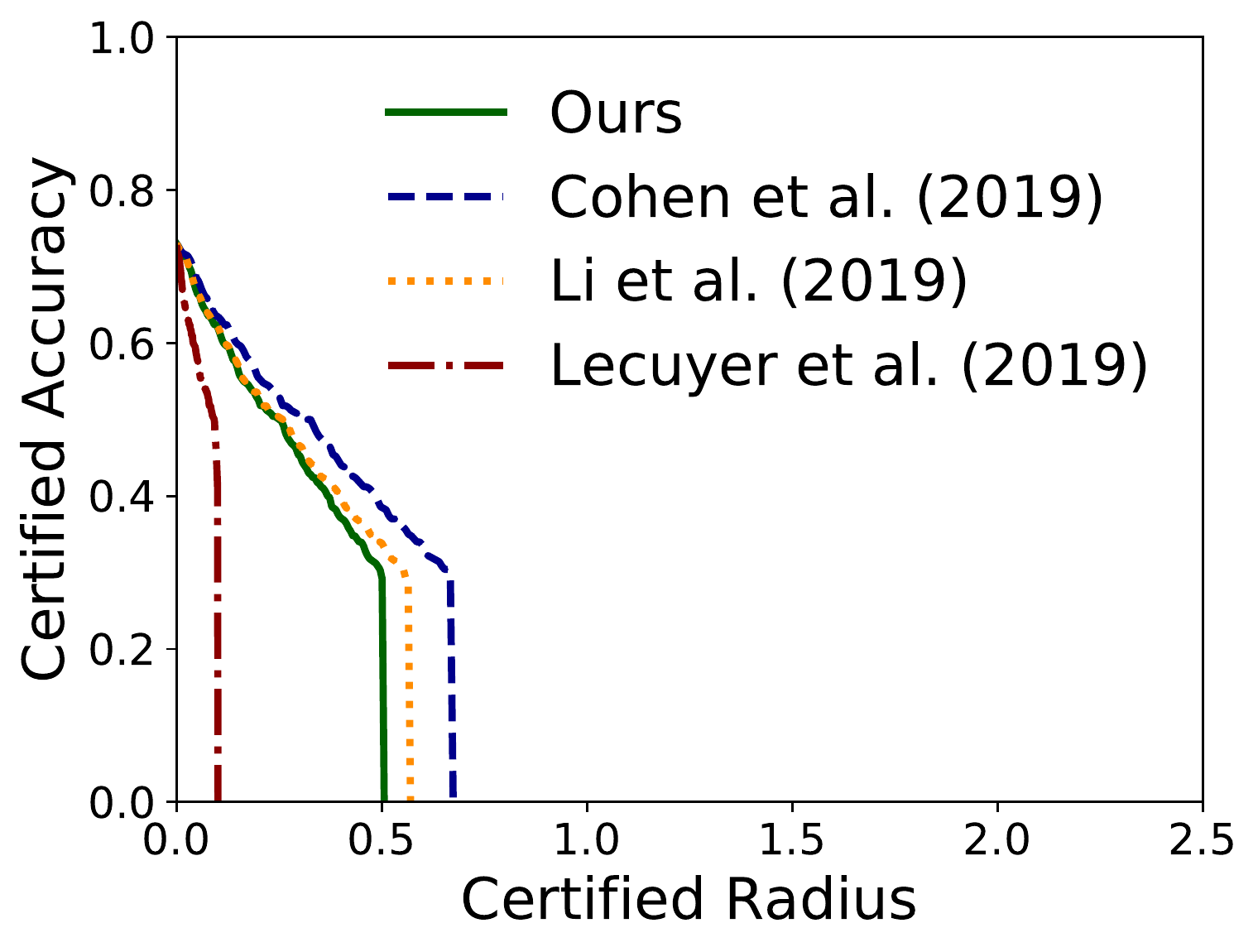}
  \caption{ImageNet, $\ell_2$}
  \label{fig:l2_cert_imagenet}
\end{subfigure}
\caption{Certified accuracy against perturbations targeting the $\ell_1$ and $\ell_2$ norms. Given as a function of the certified radius, the radius around which an input is robust.}
\label{fig:lp_cert}
\end{figure*}

\subsection{Certification guarantees beyond the $\ell_2$ based perturbations via different smoothing measures}

The Gaussian distribution is a natural choice for the smoothing measure because it naturally leads to robustness guarantees in the $\ell_2$ norm. However, it is also a convenient choice of smoothing measure because it is a member of the location-scale family of distributions. This means that, fixing $x\in\mathcal{X}$, sampling from $x + \mathcal{N}(0,\sigma^2)$ is equivalent to sampling from $\mathcal{N}(x, \sigma^2)$. Importantly, addition of a constant, $x$, does not change the family of the smoothing measure, and so we can use well known formula for the distances between two Gaussian distributions to derive robustness guarantees. Unfortunately, not all distributions belong to the location-scale family, and so, in our formulation, we are not free to choose any distribution for smoothing. Another convenient choice of a location-scale distribution is the generalized Gaussian distribution \cite{nadarajah2005generalized}, 
denoted $\mathcal{GN}(\mu, \sigma, s)$, whose density function is given
by

\begin{align}
    p(x) = \frac{s}{2\sigma\Gamma(\frac{1}{s})}e^{-|\frac{x-\mu}{\sigma}|^s} 
\end{align}

where $\mu$ is the mean, $\sigma$ denotes a scaling factor and $s$ denotes a shaping factor. The Laplacian distribution is recovered when $s=1$, the Gaussian $\mathcal{N}(\mu, \frac{\sigma^2}{2})$ when $s=2$, and the uniform distribution on $(\mu - \sigma, \mu+\sigma)$ as $s\rightarrow \infty$. We will show that by using this smoothing measure we can find robustness guarantees to $\ell_p$ perturbations, where $p \in \mathbb{N}_{>0}$.

We show in \cref{app: gn_kl_proof} that given inputs
$x$ and $x'$ the Kullback–Leibler (KL) divergence of $\mathcal{GN}(x, \sigma, s)$ and $\mathcal{GN}(x', \sigma, s)$ is given by

\begin{align}
    \sum_{k=1}^{s}{s \choose k}\frac{(1+(-1)^{s-k})\Gamma(\frac{s-k+1}{s})\norm{x - x'}_k^k}{2\sigma^k\Gamma(\frac{1}{s})} \label{eq: gn_kl}
\end{align}

We also show in \cref{app: lb_r_kl} that the
KL divergence of two multinomial
distributions $P$ and $Q$ (that disagree on the index of the top probability) is lower bounded by

\begin{align}
    d_{KL}(Q,P) \geq -\log(2\sqrt{p_1p_2} + 1 - p_1 - p_2)
\end{align}

Then we use the data processing inequality to prove robustness up to  
$\norm{x-x'}_p<\epsilon$ if the following holds

\begin{align}
    & d_{KL}(f(x + \mathcal{GN}(0, \sigma, p)), f(x' + \mathcal{GN}(0, \sigma, p))) \\[10pt]
    &\leq d_{KL}(x + \mathcal{GN}(0, \sigma, p), x' + \mathcal{GN}(0, \sigma, p)) \label{eq:gen_norm_bound1} \\[10pt]
    &\leq \frac{\epsilon^p}{\sigma^p} + \sum_{k=1}^{p-1}{p \choose k}\frac{(1+(-1)^{p-k})\Gamma(\frac{p-k+1}{p})\norm{x - x'}_k^k}{2\sigma^k\Gamma(\frac{1}{p})} \label{eq:gen_norm_bound2}\\[10pt]
    &\leq -\log(2\sqrt{p_1p_2} + 1 - p_1 - p_2) \label{eq: gen_norm_bound3}
\end{align}

\begin{table}[]
\caption{Examples of the KL divergence between $\mathcal{GN}(\mu_1, \sigma, s)$ and $\mathcal{GN}(\mu_2, \sigma, s)$ for small $s$.}
\vspace{0.3em}
\centering
\label{tab:lp_gn_examples}
\begin{tabular}{ccc}
\toprule
$s$ & $\ell_s$ & $d_{KL}(p_1,p_2)$\\
\midrule
$1$ & $\ell_1$ &
$\frac{1}{\sigma}\norm{\mu_1 - \mu_2}_1$                                                               \vspace{0.3em}                         \\

$2$ & $\ell_2$ & $\frac{1}{\sigma^2}\norm{\mu_1 - \mu_2}_2^2$                                                             \vspace{0.3em}                       \\

$3$ & $\ell_3$ & $\frac{1}{\sigma^3}\norm{\mu_1 - \mu_2}_3^3 + \frac{3}{\sigma\Gamma(\frac{1}{3})}\norm{\mu_1 - \mu_2}_1$              \vspace{0.3em}          \\

$4$ & $\ell_4$ & $\frac{1}{\sigma^4}\norm{\mu_1 - \mu_2}_4^4 + \frac{6\Gamma(\frac{3}{4})}{\sigma^2\Gamma(\frac{1}{4})}\norm{\mu_1 - \mu_2}_2^2$\\
\bottomrule
\end{tabular}
\end{table}

\Cref{tab:lp_gn_examples} gives examples of the KL-divergence of the generalized Gaussian distribution for small $\ell_p$ norms. For 
$\ell_p$ norms with $p=1$ or $p=2$,
the upper bound to which an input is certifiably robust is given by 

\begin{align}
    (-\sigma^p\log(2\sqrt{p_1p_2} + 1 - p_1 - p_2))^{\frac{1}{p}} \label{eq:l1_l2_kl_bound}
\end{align}

For $\ell_p$ norms with $p>2, p\in\mathbb{N}$, the upper bound to which an input is certifiably robust is given by $
\epsilon$ satisfying

\begin{align}
\begin{split}
\frac{\epsilon^p}{\sigma^p} + \sum_{k=1}^{p-1}{p \choose k}\frac{(1+(-1)^{p-k})\Gamma(\frac{p-k+1}{p})d^{1-\frac{k}{p}}\epsilon^k}{2\sigma^k\Gamma(\frac{1}{p})}  \\
\leq -\log(2\sqrt{p_1p_2} + 1 - p_1 - p_2) \label{eq:lp_kl_bound}
\end{split}
\end{align}

The bound given by \cref{eq:lp_kl_bound} is found by noting that $\norm{x - x'}_k \leq d^{\frac{1}{k}-\frac{1}{p}}\norm{x - x'}_p$, 
where $d$ is the dimensionality of the data. We can improve upon this naive bound to prove robustness for all norms smaller than $p$ in parallel. Without loss of generality, assume $p$ is even~\footnote{A similar statement holds when $p$ is not even.}, then we can prove robustness for every $0<k\leq p$, where $k$ is even, up to $\norm{x-x'}_k < \epsilon_k$ by solving the constrained problem

\begin{align}
\max \quad  & \epsilon_2, \epsilon_4, ..., \epsilon_p  \label{eq:lp_kl_opt_1}\\
    \textrm{subject to} \nonumber \\  
    \begin{split}& \sum_{k=1}^{p}{p \choose k}\frac{(1+(-1)^{p-k})\Gamma(\frac{p-k+1}{p})\epsilon_k^k}{2\sigma^k\Gamma(\frac{1}{p})} \\
    &\leq -\log(2\sqrt{p_1p_2} + 1 - p_1 - p_2) \label{eq:lp_kl_opt_2} 
    \end{split}
    \\
    &\epsilon_{i+2} \leq \epsilon_i \leq d^{\frac{1}{i} - \frac{1}{i+2}}\epsilon_{i+2} \label{eq:lp_kl_opt_3}\\
    &\epsilon_i > 0, \quad 2\leq i \leq p-2, \quad i\equiv 0 \Mod{2} \label{eq:lp_kl_opt_4}
\end{align}

Note that the certified radius of robustness around an input is probabilistic because we can only estimate $p_1$
and $p_2$, however, we can bound the probability of error to be arbitrarily small. In practice we follow the methods in \cite{cohen2019certified, lecuyer2018certified, li2018second} for estimating $p_1$ and $p_2$. Prediction error is bounded by collecting $n$ samples of $f(x+\rvtheta)$, where $\rvtheta$ is sampled from a generalized Gaussian distribution, and using the
Clopper-Pearson Bernoulli confidence interval to obtain a lower bound estimate of $p_1$ and an upper bound estimate of $p_2$, that holds with probability $1-\gamma$ over the
$n$ samples, where $\gamma\ll 1$. Alternatively, we can use the Hoeffding inequality which gives a lower bound of prediction error of 
$1 - ce^{-2n\epsilon^2}$, where $c$ is the number of classes $|P|$, $n$ is the number of samples and 
$\epsilon$ is the perturbation size. Clearly the error becomes arbitrarily small as we increase the number of samples.

\section{Discussion \& experiments}
\label{sec: experiments}

\begin{figure*}[t!]
\centering
\begin{subfigure}[t]{.495\textwidth}
  \centering
  \includegraphics[width=\linewidth]{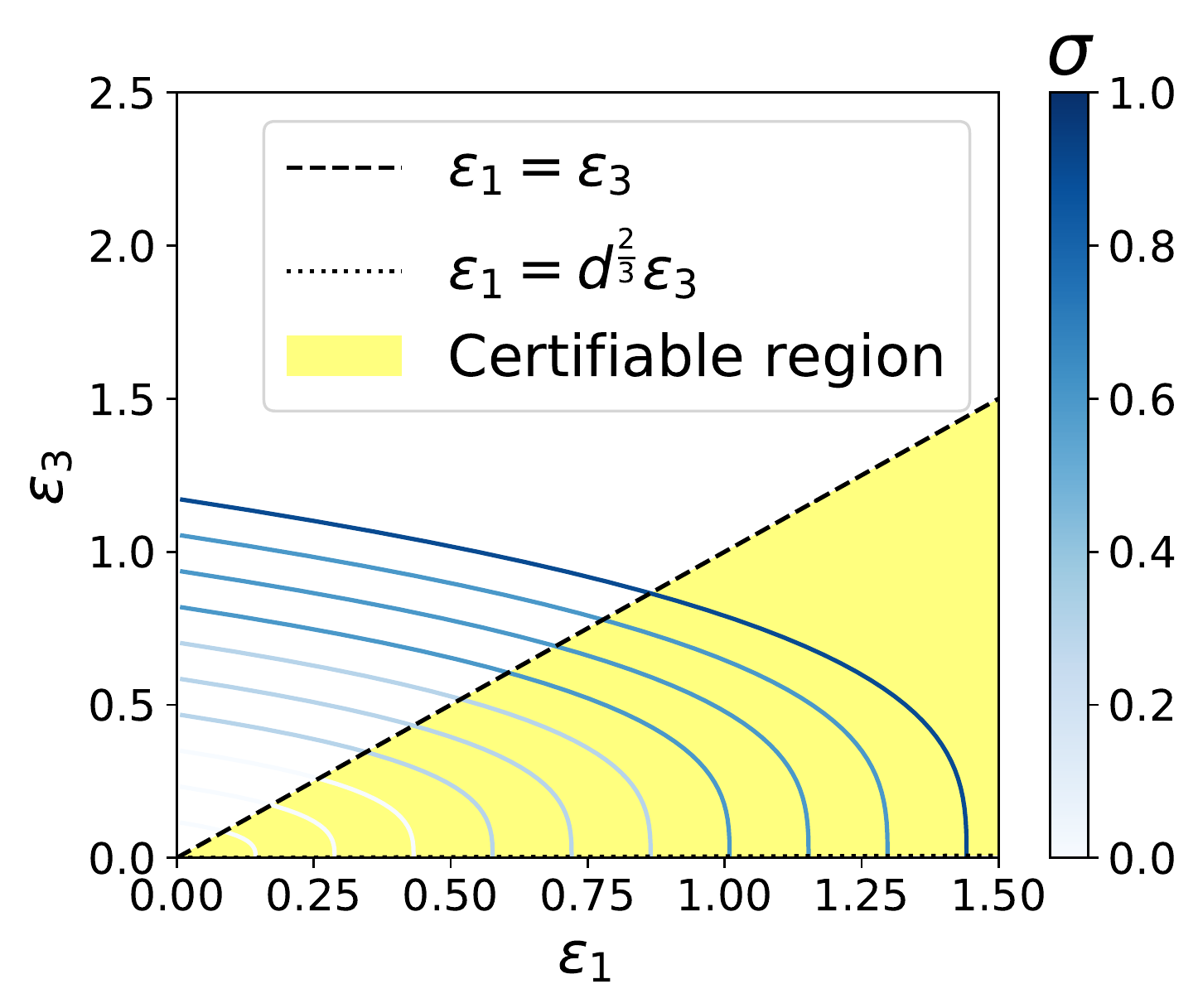}
  \caption{Certified radius trade-off between $\epsilon_3$ ($\ell_3$ norm) and $\epsilon_1$ ($\ell_1$ norm).}
  \label{fig:lp_attack_comp_l3}
\end{subfigure}
\begin{subfigure}[t]{.495\textwidth}
  \centering
  \includegraphics[width=1.0\textwidth]{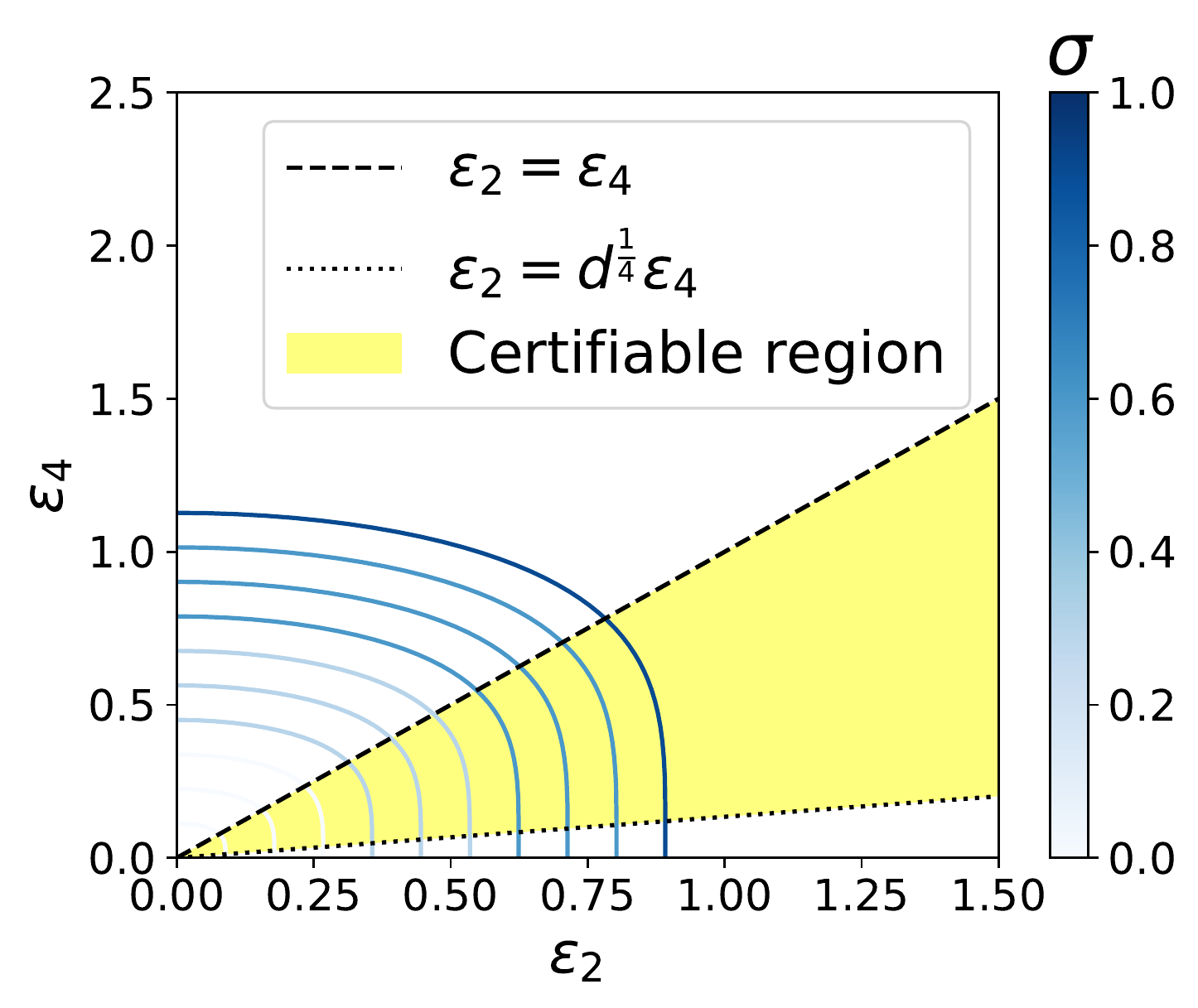}
  \caption{Certified radius trade-off between $\epsilon_4$ ($\ell_4$ norm) and $\epsilon_2$ ($\ell_2$ norm).}
  \label{fig:lp_attack_comp_l4}
\end{subfigure}
\caption{Trade-off in adversarial robustness between different norms, as we vary the noise scale, $\sigma$. We plot for a data dimensionality, $d$, equal to $3\times 32\times 32$ (the dimension for CIFAR-10 inputs), and mark the region which gives valid certificates, assuming $\hat{p_1}=0.99$ and $\hat{p_2}=1-\hat{p_1}$. }
\label{fig:lp_attack_comp}
\end{figure*}

We experimentally validated the certification procedure 
on the CIFAR-10~\cite{krizhevsky2009learning} and ImageNet~\cite{imagenet_cvpr09} datasets.
The base classifier is ResNet-50 on
ImageNet and ResNet-110 on CIFAR-10~\cite{he2016deep}.
Given an input $x$ and a classifier $f$ the certification procedure is as follows:

\begin{enumerate}
    \item Collect $n_0$ Monte Carlo samples of $f(x + \rvtheta_j)$ to estimate the true class $y$, where $\rvtheta_j \sim \mathcal{GN}(0, \sigma, s)$ and $j\in[1,...,n_0]$, with confidence $> 1-\gamma_0$.
    \item Use $n_1$ Monte Carlo samples to estimate, $\hat{p_1}$, a lower bound of the probability of the most-likely class with confidence $>1-\gamma_1$. We follow Cohen \etal \cite{cohen2019certified} for estimating $ \hat{p_2}$, an upper bound of the probability of the second most-likely class, who noticed nearly all  probability mass on other classes is placed on the second most-likely class and so use $ \hat{p_2} = 1 - \hat{p_1}$.
    \item Use $\hat{p_1}$, $\hat{p_2}$ and \cref{eq:l1_l2_kl_bound} or \cref{eq:lp_kl_bound}  to find a certified radius around $x$.
\end{enumerate}

For all experiments we use $n_0=100, n_1=100,000, \gamma_{\{0,1\}}=0.001, \sigma=0.25$ and certify 400 test set examples for both CIFAR-10 and ImageNet datasets~\footnote{We perform experiments measuring the effect that various $\sigma$ have on the certified radius in \cref{app: sigma_certify}.}. Theoretically, 
 this procedure can certify any classifier, however in practice, image classifiers are not stable under noise and so we found it necessary to train classifiers with generalized Gaussian noise (using the same scale and shape parameters as is used during certification). Note that this has the same complexity as standard data augmentation during training and is less expensive than the Madry et al. \cite{madry2017towards} defense. 

\subsection{Comparison to related work}

For both CIFAR-10 and ImageNet we certify inputs against perturbations in $\ell_1$ and $\ell_2$ norms and compare against \cite{cohen2019certified, lecuyer2018certified, li2018second}.
~\Cref{fig:lp_cert} shows 
certified accuracy as a function
of the certified radius. 
In general, the largest certified regions come against perturbations targeting the $\ell_1$ norm. In \cref{app: viz_samples}, we show qualitative examples of inputs smoothed with generalized Gaussian noise and the corresponding robustness guarantees in the $\ell_1$, $\ell_2$, and $\ell_3$ norms.

While the primary boon of our certification procedure is its ability to certify inputs to adversarial perturbations beyond
$\ell_1$ and $\ell_2$ norms, the method is not substantially weaker than related work in either norm. In
\cref{fig:l1_cert_cifar10} and \cref{fig:l1_cert_imagenet}, we compare with
Lecuyer \etal \cite{lecuyer2018certified} and Li \etal  \cite{li2018second}  for $\ell_1$ norm certificates. Given estimates $\hat{p_1}$ and
$\hat{p_2}$, Lecuyer \etal \cite{lecuyer2018certified} find a certified radius against $\ell_1$ perturbations given by $\frac{\sigma}{2} \log(\hat{p_1} / \hat{p_2})$, while Li \etal \cite{li2018second} find a certified radius against $\ell_1$ perturbations given by $\sigma \log(1 - \hat{p_1} + \hat{p_2})$. Li \etal \cite{li2018second} and Teng \etal \cite{teng2020ell} show that this robustness guarantee is tight for the $\ell_1$ norm. Our $\ell_1$ certificates are slightly weaker than Lecuyer \etal \cite{lecuyer2018certified}, and both are dominated by Li \etal \cite{li2018second} who obtain the tightest possible certificates.

In \cref{fig:l2_cert_cifar10} and \cref{fig:l2_cert_imagenet}, we compare with
Lecuyer \etal \cite{lecuyer2018certified}, Li \etal \cite{li2018second}, and Cohen \etal \cite{cohen2019certified} for $\ell_2$ norm certificates. 
Our $\ell_2$ certificates strictly dominate Lecuyer \etal \cite{lecuyer2018certified}, and are approximately equivalent to Li \etal \cite{li2018second}. This equivalence is to be expected since our certificates are closely related to Li \etal \cite{li2018second} certificates, which are based on the Rényi divergence between two Gaussians, while ours are based on KL divergence. Clearly, we could improve upon this $\ell_2$ guarantee if we used the chi-squared distance instead of KL divergence and a standard Gaussian smoothing measure, as proved by \cref{prop_main}. However, our aim is to show the general capacity of the generalized Gaussian as a smoothing measure for certification.

\begin{figure*}[t!]
\centering
        \begin{subfigure}[t]{0.339\textwidth}
                \includegraphics[width=\linewidth]{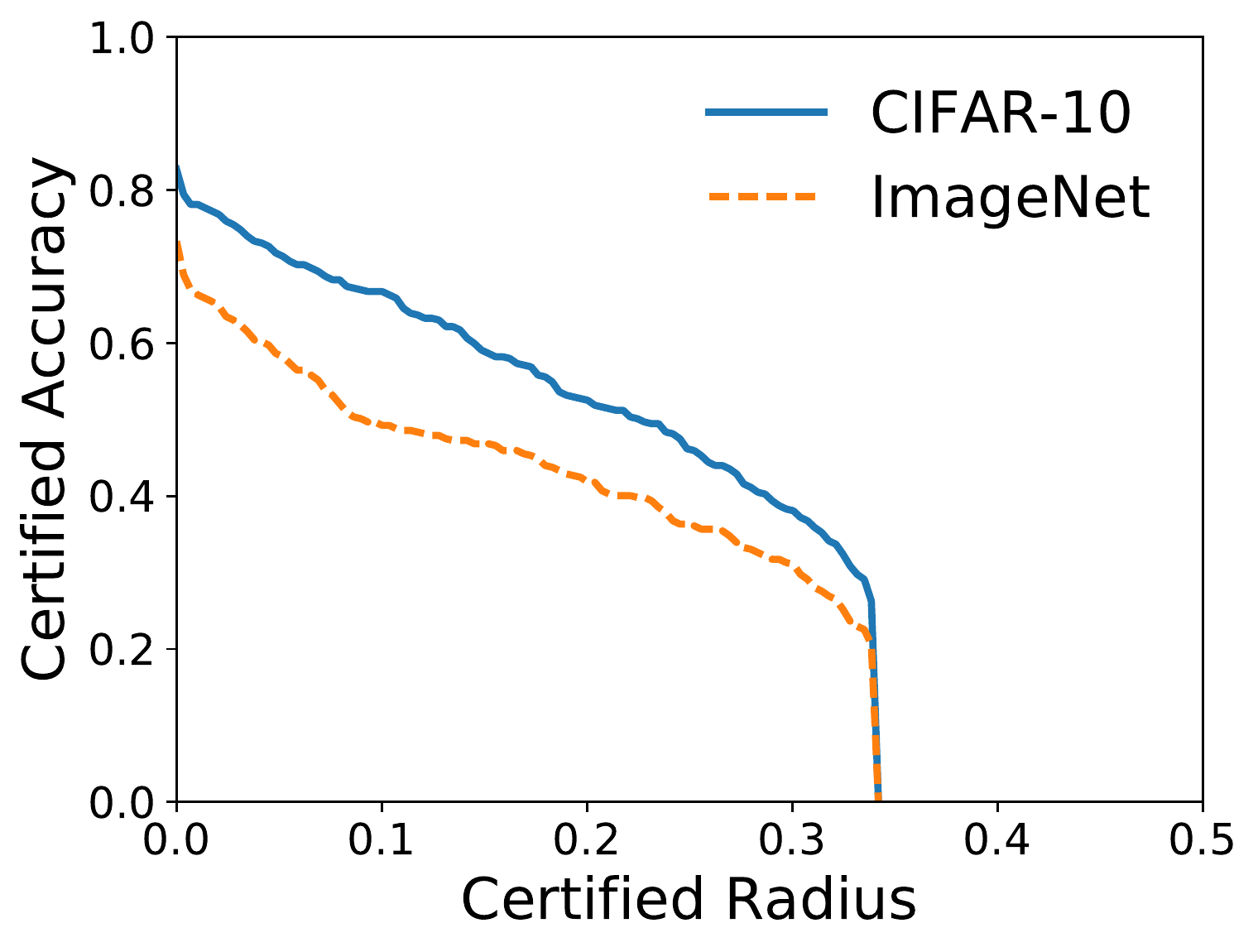}
                \caption{$\ell_3$}
                \label{fig:gull}
        \end{subfigure}%
        \begin{subfigure}[t]{0.339\textwidth}
                \includegraphics[width=\linewidth]{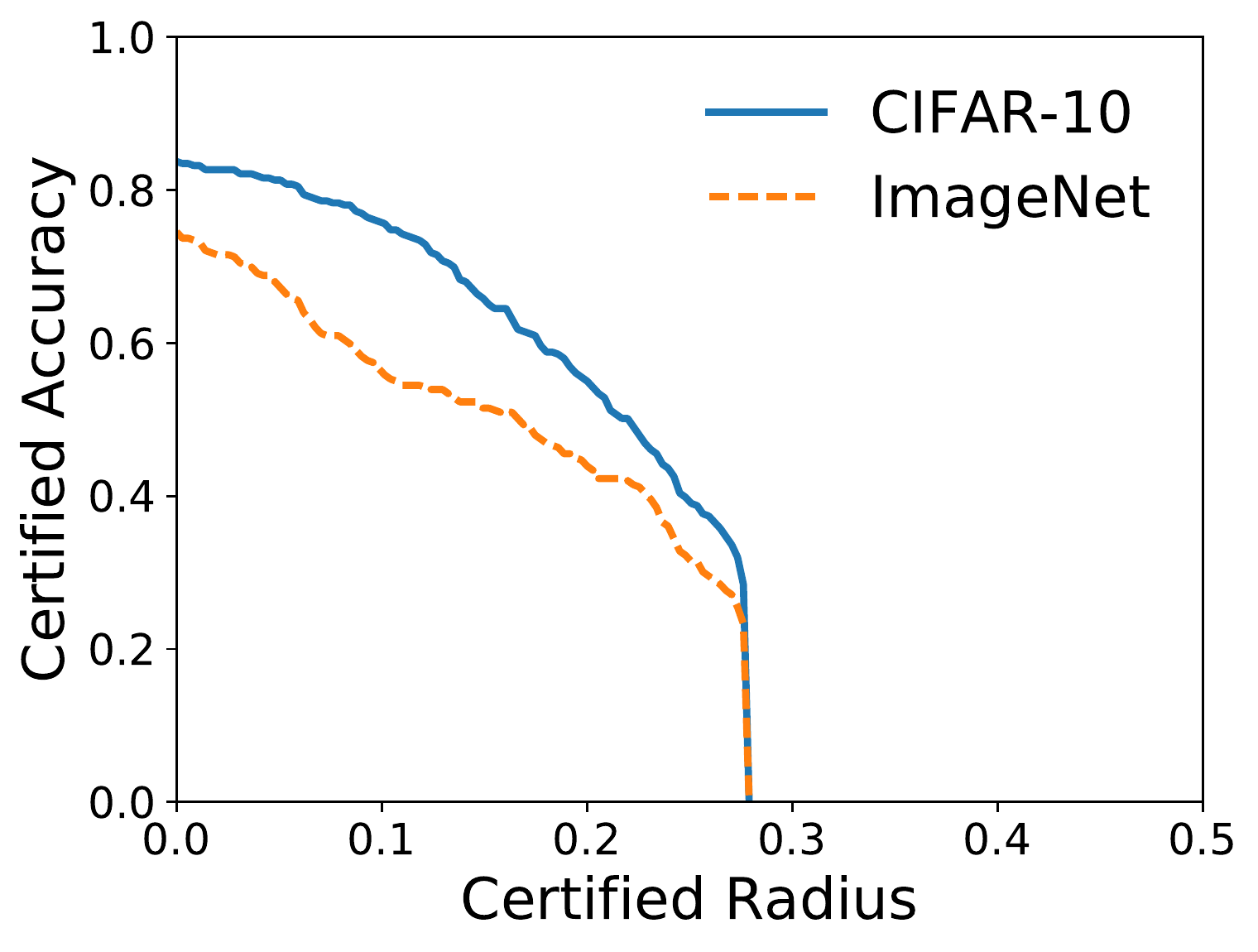}
                \caption{$\ell_4$}
                \label{fig:gull2}
        \end{subfigure}%
        \begin{subfigure}[t]{0.339\textwidth}
                \includegraphics[width=\linewidth]{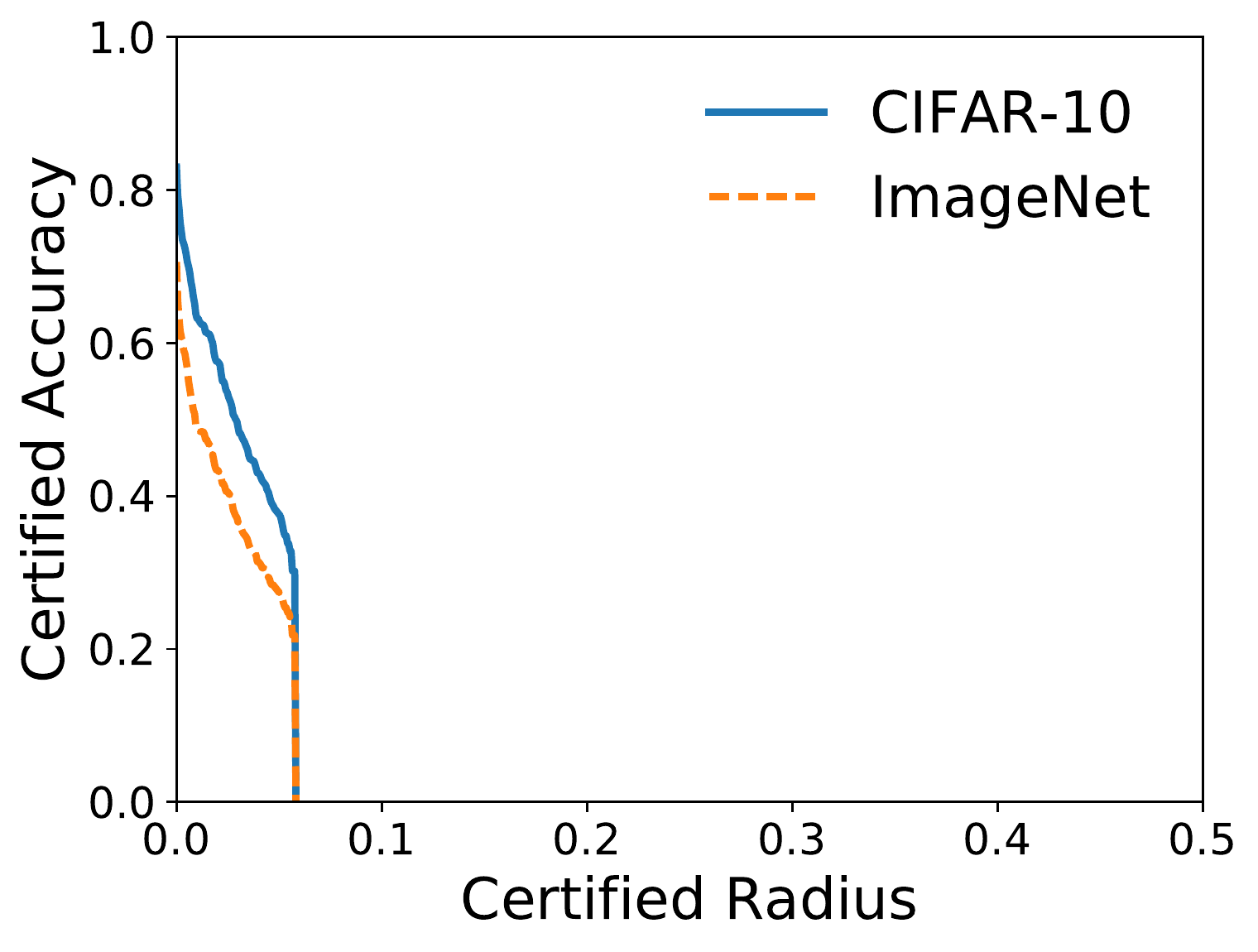}
                \caption{$\ell_5$}
                \label{fig:tiger}
        \end{subfigure}%
\caption{Certified accuracy on 400 CIFAR-10 test set inputs and 400 ImageNet test set inputs against perturbations targeting the $\ell_3$, $\ell_4$, and $\ell_5$  norms. Given as a function of the certified radius, the radius around which an input is robust. Inputs were smoothed under a generalized Gaussian distribution parameterized by $\mathcal{GN}(0, 0.25, p)$.}
\label{fig: lp_radius}
\end{figure*}

\subsection{Robustness trade-offs between different $\ell_p$ norms.}

As described by \cref{eq:lp_kl_bound}, to obtain robustness guarantees in $\ell_{p>2}$ norms we must factor in required robustness guarantees in smaller $\ell_p$ norms. For example, to prove robustness up to $\norm{x-x'}_3 <\epsilon_3$ and $\norm{x-x'}_1 <\epsilon_1$ we find $\epsilon_1$ and $\epsilon_3$ satisfying 


\begin{align}
\begin{split}
&\frac{1}{\sigma^3}\epsilon_3^3 + \frac{3}{\sigma\Gamma(\frac{1}{3})}\epsilon_1
\leq -\log(2\sqrt{\hat{p_1}\hat{p_2}} + 1 - \hat{p_1} - \hat{p_2}) \\&\wedge \label{eq:l3_prob}\\ &0<\epsilon_3\leq\epsilon_1\leq d^{\frac{2}{3}}\epsilon_3, 
\end{split}
\end{align}


and to prove robustness up to $\norm{x-x'}_4 <\epsilon_4$ and $\norm{x-x'}_2 <\epsilon_2$ we find $\epsilon_2$ and $\epsilon_4$ satisfying 

\begin{align}
\begin{split}
&\frac{1}{\sigma^4}\epsilon_4^4 + \frac{6\Gamma(\frac{3}{4})}{\sigma^2\Gamma(\frac{1}{4})}\epsilon_2^2
\leq -\log(2\sqrt{\hat{p_1}\hat{p_2}} + 1 - \hat{p_1} - \hat{p_2})\\ &\wedge \label{eq:l4_prob}\\ &0<\epsilon_4\leq\epsilon_2\leq d^{\frac{1}{4}}\epsilon_4,
\end{split}
\end{align}

We visualize this trade-off in
\cref{fig:lp_attack_comp} for $\ell_3$ and $\ell_4$ norms. That is, the
trade-off in certified robustness between those norms and certified robustness
in $\ell_1$ and $\ell_2$, respectively. We visualize the trade-off as we vary the noise scale $\sigma$, assuming a robust classifier that classifies inputs correctly with $\hat{p_1}=0.99$ and $\hat{p_2}=0.01$. We can smoothly exchange robustness in one norm for robustness in another norm. For example, given $\sigma=1$ and a CIFAR-10 input,
we can reduce the guaranteed robustness in the $\ell_3$
norm from an approximate certified radius of 0.86 to approximately 0, and increase the guaranteed robustness in the  $\ell_1$
norm from a certified radius of 0.86 to 1.44.
In \cref{fig: lp_radius}, we show certified accuracy as a function of certified radius in the $\ell_3$, $\ell_4$, and $\ell_5$ norms on the CIFAR-10 and ImageNet datasets. To find the maximum $\epsilon_3$ we solve \cref{eq:l3_prob} such that $\epsilon_3 = \epsilon_1$. Similarly for $\epsilon_4$ we solve \cref{eq:l4_prob} such that $\epsilon_4 = \epsilon_2$, and extend this line of reasoning to find $\epsilon_5 = \epsilon_3 = \epsilon_1$ for the $\ell_5$ norm. Clearly, we can find non-negligible certified radii in norms outside of $\ell_1$ and $\ell_2$.

\subsection{Robustness guarantees as $\ell_{p\rightarrow \infty}$.}

An immediate question arises when observing our certification procedure, can we find non-vacuous robustness guarantees for arbitrarily large $\ell_p$ norms, where $p$ is even~\footnote{Equivalent results for this section can be found when $p$ is not even.}~\footnote{The subject of simultaneous robustness over every $\ell_p$ norm is expanded upon in \cref{sec: lp_sphere_example}.}? Given \cref{eq:lp_kl_opt_2}, note that $\nicefrac{{p \choose k}(1+(-1)^{p-k})\Gamma(\frac{p-k+1}{p})}{2\Gamma(\frac{1}{p})}\geq 1$, $\forall 1\leq k\leq p$, where $k$ is even, and as $p\rightarrow \infty$, $\exists k$ such that $\nicefrac{{p \choose k}(1+(-1)^{p-k})\Gamma(\frac{p-k+1}{p})}{2\Gamma(\frac{1}{p})}\rightarrow \infty$. We must therefore solve the problem given in \cref{eq:lp_kl_opt_1}-\cref{eq:lp_kl_opt_4}, where \cref{eq:lp_kl_opt_2} is given by


\begin{align}
    &\frac{c_2\epsilon_2^2}{\sigma^2} + \frac{c_4\epsilon_4^4}{\sigma^4} + ... + \frac{c_p\epsilon_p^p}{\sigma^p} \leq -\log(2\sqrt{p_1 p_2} + 1 - p_1 - p_2) \label{eq:new_lp_kl_opt_2} \\ 
    &\text{ where } c_k\in \mathbb{R}_{>1}, 1\leq k \leq p, k\equiv 0 \Mod{2} 
\end{align}

To satisfy \cref{eq:lp_kl_opt_3}, we can find $\epsilon_2, \epsilon_4, ..., \epsilon_p$ such that $\epsilon_2= \epsilon_4= ...= \epsilon_p$; we refer to this value as $\epsilon$, and \cref{eq:new_lp_kl_opt_2} becomes


\begin{align}
\begin{split}
    &c_2(\frac{\epsilon}{\sigma})^2 + c_4(\frac{\epsilon}{\sigma})^4 + ... + c_p(\frac{\epsilon}{\sigma})^p \\ &\leq -\log(2\sqrt{p_1 p_2} + 1 - p_1 - p_2)\\  \label{eq:newer_lp_kl_opt_2}
\end{split}\\
&\text{ where } c_k\in \mathbb{R}_{>1}, 1\leq k \leq p, k\equiv 0 \Mod{2}
\end{align}

For a fixed $p_1, p_2, \sigma$, since $\forall k, c_k\geq 1$, and $\exists k$ such that $c_k\rightarrow \infty$ when $p\rightarrow \infty$, to satisfy the inequality in  \cref{eq:newer_lp_kl_opt_2}, we must have $\epsilon \rightarrow 0$. If we do not fix $\sigma$ then we require $(\frac{\epsilon}{\sigma})^k \rightarrow 0$ as $c_k \rightarrow \infty$, and so to certify a non-negligible radius, $\epsilon$, we require $\sigma \rightarrow \infty$. However, as $\sigma \rightarrow \infty$, the randomized smoothing will cause the input to become too noisy for any classifier to achieve low prediction error.

Clearly, as $p$ grows the largest possible certified radius becomes smaller, because our bound requires this robustness guarantee holds for every norm smaller than $p$. One may wonder if we can find an $\ell_p$ norm in which we can certify a non-vacuous radius that approximates the $\ell_{\infty}$ norm arbitrarily well. The difference in volume between a unit ball in the $\ell_p$ norm and $\ell_{\infty}$ norm is given by $\nicefrac{\Gamma(1+\nicefrac{1}{p})^d}{\Gamma(1+\nicefrac{d}{p})}$, where $d$ is the data dimensionality. Unfortunately, the error in the  approximation is dependent on the data  dimensionality. For example, for an ImageNet input where $d=3\times 224 \times 224$, if we require the ratio of volumes between an $\ell_p$ unit ball and $\ell_{\infty}$ unit ball to be larger than 0.99, we must take $p=9\times 3\times 224 \times 224 $.

\subsection{How tight is the bound?}

\begin{figure}[t!]
\centering
  \includegraphics[width=1.0\columnwidth]{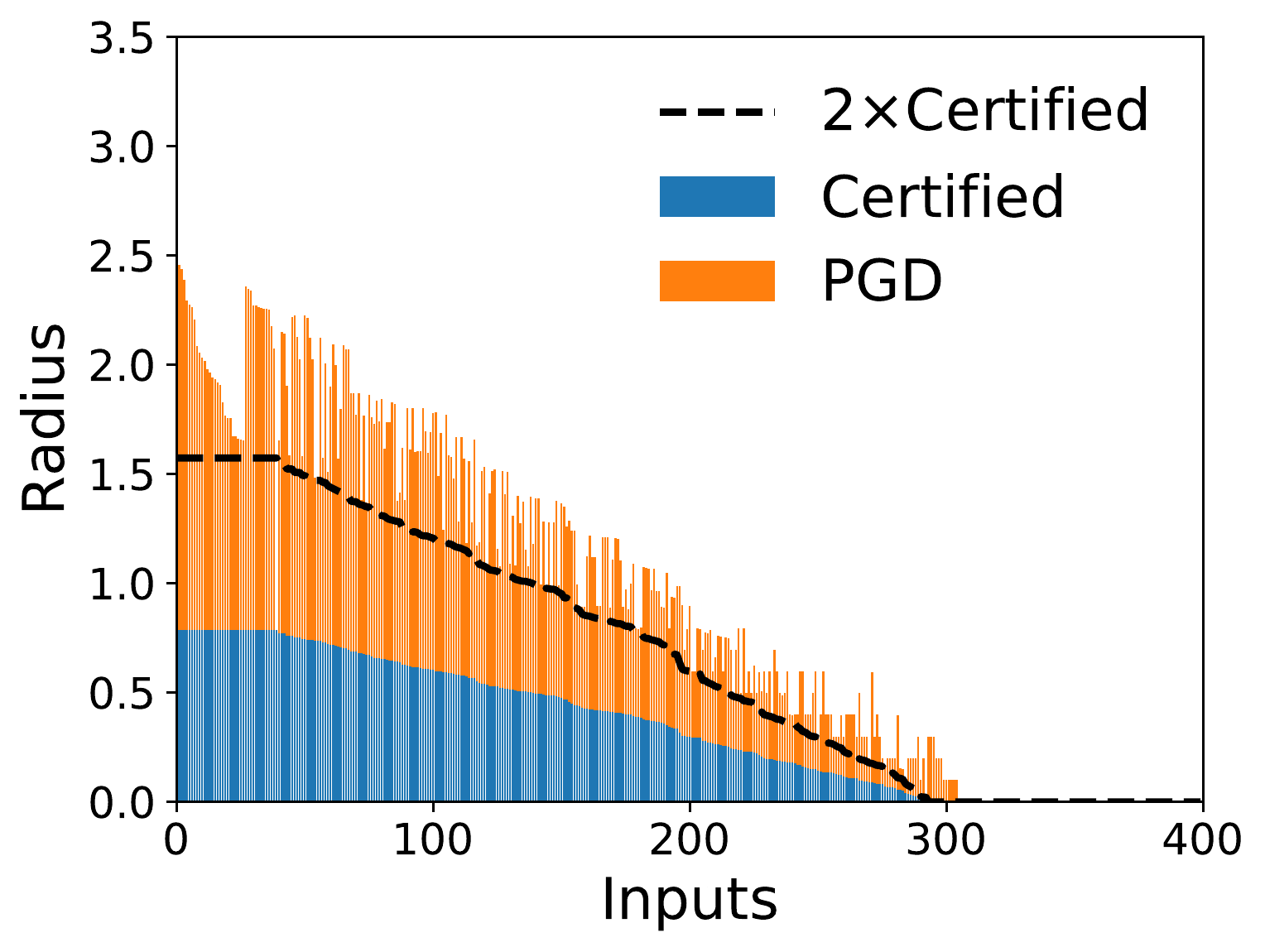}
\caption{The certified radius and size of adversarial perturbations for 400 CIFAR-10 test inputs using a PGD attack optimizing the $\ell_2$ norm. As a guide to assess how close the certified radius is to adversarial perturbation size, we also display $2\times$ the certified radius of an input.}
\label{fig: lp_attack_comp}
\end{figure}

The difference between the certified area and the size of an adversarial perturbation gives a tightness estimate. If the certified radius is close to the size of an adversarial perturbation this implies the bound is close to optimal. To check how tight our bound is we ran the PGD attack~\cite{madry2017towards} minimizing 
perturbations in the $\ell_2$ norm. Because the certification procedure requires the addition of generalized Gaussian noise to the input, the gradient is highly stochastic, leading to extremely slow convergence of the PGD attack. We circumvent this stochasticity by optimizing using the Expectation Over Transformation~\cite{athalye2017synthesizing} -- we use 1000 Monte Carlo samples to estimate the gradient of an input during the attack.~\Cref{fig: lp_attack_comp} gives attack results on CIFAR-10 along with the certified radius of 400 inputs. We find adversarial examples with norms within $2-2.5\times$ the certified radius. Unfortunately, this does not inform us if our bound is loose or if the attack is sub-optimal. We leave a more rigorous investigation of assessing the tightness of our bound for future work.

\section{Conclusion}
\label{sec: conclusion}

Randomized smoothing has offered a promising approach to scaling robustness guarantees to large architectures.
By extending the framework developed by Li \etal \cite{li2018second}, we showed how different choices of divergences affects the certified radius of robustness around an input. We verified that Rényi divergence is superior to other common f-divergences in this framework,  for certifying an input against $\ell_2$ perturbations. We then showed that a generalized Gaussian smoothing measure leads to robustness guarantees against any $\ell_p$ ($p\in\mathbb{N}_{>0}$) minimized adversarial perturbation, however, non-negligible certified radii are only available for small $\ell_p$ norms.

\section*{Acknowledgements}

Jamie Hayes is funded by a Google PhD Fellowship in Machine Learning.

{\small
\bibliographystyle{ieee}
\bibliography{references}
}

\newpage
\begin{appendix}
\onecolumn

\section{Lower bounds for common divergences between multinomial distributions}
\label{app: lb_r_kl}

Firstly, we present the statement of the Rényi divergence bound given in Li \etal \cite{li2018second}, and provide a full proof.

\begin{theorem}
\label{renyi_bound_theorem}
Let $P = (p_1,..., p_k)$ and $Q = (q_1,..., q_k)$ be two multinomial distributions over the
same index set $\{1,..., k\}$. If the indexes of the largest probabilities do not match on $P$ and $Q$, that is
$\argmax_i p_i \neq \argmax_j q_j$, then

\begin{align}
\begin{split}
d_{\alpha}(Q,P) \geq -\log\Big(&1-p_1-p_2 +\\ 
&2\big(\frac{1}{2}(p_1^{1-\alpha} + p_2^{1-\alpha})\big)^{\frac{1}{1-\alpha}}\Big)
\end{split}
\end{align}

where $p_1$ and $p_2$ are the first and second largest probabilities in $P$.
\end{theorem}

\begin{proof}
We can think of this problem as finding the
distribution Q that minimizes $d_{\alpha}(Q,P)$ such that $\argmax_i p_i \neq \argmax_j q_j$ for a fixed $P = (p_1,..., p_k)$. Without loss of generality, assume
$p_1 \geq p_2 \geq ... \geq p_k$.

This is equivalent to solving

\begin{align}
\min_{\sum q_i, \argmax q_i\neq 1} \frac{1}{1-\alpha}\log\Big(\sum_1^{k} p_i\big(\frac{q_i}{p_i}\big)^{\alpha}\Big)
\end{align}

As the logarithm is a monotonically increasing function, we only focus on the quantity $s(Q,P)= \sum_1^{k} p_i(\frac{q_i}{p_i})^{\alpha}$ for a fixed $\alpha$.

We first show for the $Q$ that minimizes $s(Q,P)$, it must have $q_1=q_2\geq q_3 \geq ... \geq q_k$.  Note here
we allow a tie, because we can always let $q_1 = q_1-\kappa$ and $q_2 = q_2 + \kappa$ for some small $\kappa$ to satisfy
$\argmax q_i \neq 1$ while not changing the Rényi divergence too much by the continuity of $s(Q,P)$.

If $q_j > q_i$ for some $j \geq i$, we can define $Q'$ by mutating $q_i$ and $q_j$, that is $Q' =
(q_1,..., q_{i - 1}, q_j , q_{i+1},..., q_{j - 1}, q_i, q_{j+1},..., q_k)$, then

\begin{align}
S(Q,P)- S(Q',P) &= p_i\big(\frac{q_i^{\alpha} - q_j^{\alpha}}{p_i^{\alpha}}\big) + p_j\big(\frac{q_j^{\alpha} - q_i^{\alpha}}{p_j^{\alpha}}\big) \\
&= (p_i^{1-\alpha} - p_j^{1-\alpha})(q_i^{\alpha} - q_j^{\alpha}) > 0
\end{align}

which conflicts with the assumption that $Q$ minimizes $s(Q,P)$. Thus $q_i \geq q_j$ for $j \geq i$. Since $q_1$
cannot be the largest, we have $q_1 = q_2 \geq q_3 \geq ... \geq q_k$.

Then we are able to assume $Q = (q_0, q_0, q_3,..., q_k)$, and the problem can be formulated as

\begin{align}
    & \min_{q_0, q_3, ..., q_k} p_1(
    \frac{q_0}{p_1})^{\alpha} + p_2(
    \frac{q_0}{p_2})^{\alpha} + \sum_{i=3}^k p_i(
    \frac{q_i}{p_i})^{\alpha} \\
    & \text{such that}\quad 2q_0 + \sum_{i=3}^k q_i = 1 \\
    & \text{such that}\quad q_i - q_0 \leq 0 \quad i\geq 3\\
    & \text{such that}\quad - q_i \leq 0 \quad i\geq 0
\end{align}

which forms a set of KKT conditions. Let $L$ denote the Lagrangian formulation of the problem

\begin{align}
    &p_1(\frac{q_0}{p_1})^{\alpha} + p_2(
    \frac{q_0}{p_2})^{\alpha} +  \sum_{i=3}^k p_i(
    \frac{q_i}{p_i})^{\alpha} + \lambda(2q_0 + \sum_{i=3}^k q_i - 1) + \sum_{i=3}^k \mu_i (q_i - q_0) - \sum_{i=3}^k \beta_i q_i
\end{align}

Setting slack variables to zero and differentiating gives

\begin{align}
    & \frac{\partial L}{\partial q_0} = \alpha q_0^{\alpha - 1}(p_1^{1-\alpha} + p_2^{1-\alpha}) + 2\lambda = 0 \label{eq:renyi_1}\\
    & \frac{\partial L}{\partial q_i} = \alpha(\frac{q_i}{p_i})^{\alpha - 1} + \lambda = 0 \quad i \geq 3 \label{eq:renyi_2}
\end{align}

\Cref{eq:renyi_1} and \cref{eq:renyi_2} imply

\begin{align}
    & q_0 = \Big( \frac{-2\lambda}{\alpha(p_1^{1-\alpha} + p_2^{1-\alpha})}\Big)^{\frac{1}{\alpha-1}}  \label{eq:renyi_3}\\
    & q_i = \Big(-\frac{\lambda}{\alpha}\Big)^{\frac{1}{\alpha-1}}p_i \quad i \geq 3 \label{eq:renyi_4}
\end{align}

From the restriction that $2q_0 + \sum_{i=3}^k q_i = 1$ it follows that

\begin{align}
    & \lambda = \frac{-\alpha}{\Big(2\big(\frac{1}{2}(p_1^{1-\alpha} + p_2^{1-\alpha})\big)^{\frac{1}{1-\alpha}} + 1 - p_1 - p_2\Big)^{\alpha -1}}  \label{eq:renyi_5}
\end{align}

Let $\eta = \big(\frac{p_1^{1-\alpha} + p_2^{1-\alpha}}{2}\big)^\frac{1}{1-\alpha}$. Then it follows that

\begin{align}
    & q_0 = \frac{a}{2\eta + 1-p_1-p_2}   \label{eq:renyi_6}\\
    & q_i = \frac{p_i}{2\eta-1-p_1-p_2} \quad i \geq 3 \label{eq:renyi_7}
\end{align}

Using \cref{eq:renyi_6} and \cref{eq:renyi_7}, Rényi divergence is minimized at

\begin{align}
    &\quad \frac{1}{1-\alpha}\log\Big(p_1(\frac{q_0}{p_1})^{\alpha} + p_2(
    \frac{q_0}{p_2})^{\alpha} + \sum_{i=3}^k p_i(
    \frac{q_i}{p_i})^{\alpha}\Big)\\
    &= \frac{1}{1-\alpha}\log\Big(\frac{2\eta^{1-\alpha}\eta^{\alpha}}{(2\eta + 1 -p_1-p_2)^{\alpha}} + \frac{1-p_1-p_2}{(2\eta + 1 -p_1-p_2)^{\alpha}}\Big)\\
    &= -\log(2\eta + 1 - p_1 - p_2)
\end{align}

\end{proof}

To find the certified area of robustness of an input using the KL divergence of the generalized Gaussian norm, we can make use of the following theorem.

\begin{theorem}
\label{kl_bound_theorem}
Let $P = (p_1,..., p_k)$ and $Q = (q_1,..., q_k)$ be two multinomial distributions over the
same index set ${1,..., k}$. If the indexes of the largest probabilities do not match on $P$ and $Q$, that is
$\argmax_i p_i \neq \argmax_j q_j$, then

\begin{align}
d_{KL}(Q,P) \geq -\log(2\sqrt{p_1p_2} + 1 - p_1 - p_2)
\end{align}

where $p_1$ and $p_2$ are the first and second largest probabilities in $P$.
\end{theorem}

\begin{proof}
Using the same terminology as \Cref{renyi_bound_theorem}, the problem can be 
stated as a set of KKT conditions given by

\begin{align}
    & \min_{q_0, q_3, ..., q_k} q_0
    \log(\frac{q_0}{p_1}) + q_0\log(
    \frac{q_0}{p_2}) + \sum_{i=3}^k q_i\log(
    \frac{q_i}{p_i})\\
    & \text{such that}\quad 2q_0 + \sum_{i=3}^k q_i = 1 \\
    & \text{such that}\quad q_i - q_0 \leq 0 \quad i\geq 3\\
    & \text{such that}\quad - q_i \leq 0 \quad i\geq 0
\end{align}

Let $L$ denote 

\begin{align}
    &p_1\log(\frac{q_0}{p_1}) + p_2\log(
    \frac{q_0}{p_2}) + \sum_{i=3}^k p_i\log(
    \frac{q_i}{p_i}) +\\
    &\lambda(2q_0 + \sum_{i=3}^k q_i - 1) + \sum_{i=3}^k \mu_i (q_i - q_0) - \sum_{i=3}^k \beta_i q_i
\end{align}

Setting slack variables to zero and differentiating gives

\begin{align}
    & \frac{\partial L}{\partial q_0} = \log(\frac{q_0}{p_1}) + \log(\frac{q_0}{p_2}) + 2\lambda + 2 = 0 \label{eq:kl_1}\\
    & \frac{\partial L}{\partial q_i} = \log(\frac{q_i}{p_i}) + \lambda +1=0 \quad i \geq 3 \label{eq:kl_2}
\end{align}

Combining \cref{eq:kl_1} and \cref{eq:kl_2} with
the KKT conditions and solving for $\lambda$ gives

\begin{align}
    &q_0 =\frac{\sqrt{p_1p_2}}{\eta}
    \\
    &q_i = \frac{p_i}{\eta} \quad\quad i \geq 3 \\
\end{align}

where $\eta=2\sqrt{p_1p_2} + 1 - p_1 - p_2$. The minimized KL divergence is therefore $-\log\eta$.

\end{proof}

\begin{theorem}
\label{hellinger_bound_theorem}
Let $P = (p_1,..., p_k)$ and $Q = (q_1,..., q_k)$ be two multinomial distributions over the
same index set ${1,..., k}$. If the indexes of the largest probabilities do not match on $P$ and $Q$, that is
$\argmax_i p_i \neq \argmax_j q_j$, then

\begin{align}
d_{H^2}(Q,P) \geq 1-\sqrt{\frac{2-(\sqrt{p_1}-\sqrt{p_2})^2}{2}}
\end{align}

where $p_1$ and $p_2$ are the first and second largest probabilities in $P$.
\end{theorem}

\begin{proof}

Using the same technique and terminology as in \Cref{renyi_bound_theorem}, we find that

\begin{align}
    &q_0 =\frac{(\sqrt{p_1}+\sqrt{p_2})^2}{2\eta}
    \\
    &q_i = \frac{2p_i}{\eta} \quad\quad i \geq 3, \\
\end{align}

where $\eta=2-(\sqrt{p_1}-\sqrt{p_2})^2$. The minimized Hellinger distance is therefore $1-\sqrt{\frac{\eta}{2}}$.

\end{proof}

\begin{theorem}
\label{chi_squared_bound_theorem}
Let $P = (p_1,..., p_k)$ and $Q = (q_1,..., q_k)$ be two multinomial distributions over the
same index set ${1,..., k}$. If the indexes of the largest probabilities do not match on $P$ and $Q$, that is
$\argmax_i p_i \neq \argmax_j q_j$, then

\begin{align}
d_{\chi^2}(Q,P) \geq \frac{(p_1-p_2)^2}{(p_1 + p_2) - (p_1 -p_2)^2}
\end{align}

where $p_1$ and $p_2$ are the first and second largest probabilities in $P$.
\end{theorem}

\begin{proof}

Using the same technique and terminology as in \Cref{renyi_bound_theorem}, we find that

\begin{align}
    &q_0 =\frac{2p_1p_2}{\eta}
    \\
    &q_i = \frac{p_1 + p_2}{\eta}p_i \quad\quad i \geq 3, \\
\end{align}

where $\eta=(p_1 + p_2) - (p_1 -p_2)^2$. The minimized chi-squared distance is therefore $\frac{(p_1-p_2)^2}{\eta}$.

\end{proof}

\begin{theorem}
\label{bhat_bound_theorem}
Let $P = (p_1,..., p_k)$ and $Q = (q_1,..., q_k)$ be two multinomial distributions over the
same index set ${1,..., k}$. If the indexes of the largest probabilities do not match on $P$ and $Q$, that is
$\argmax_i p_i \neq \argmax_j q_j$, then

\begin{align}
d_{B}(Q,P) \geq -\log(\sqrt{\frac{2\sqrt{p_1p_2} -p_1-p_2 + 2}{2}})
\end{align}

where $p_1$ and $p_2$ are the first and second largest probabilities in $P$.
\end{theorem}

\begin{proof}

Using the same technique and terminology as in \Cref{renyi_bound_theorem}, we find that

\begin{align}
    &q_0 =\frac{(\sqrt{p_1} + \sqrt{p_2})^2}{2\eta}
    \\
    &q_i = \frac{2p_i}{\eta} \quad\quad i \geq 3, \\
\end{align}

where $\eta=2\sqrt{p_1p_2} -p_1-p_2 + 2$. The minimized Bhattacharyya distance is therefore $-log(\sqrt{\frac{\eta}{2}})$.

\end{proof}

\begin{theorem}
\label{tv_bound_theorem}
Let $P = (p_1,..., p_k)$ and $Q = (q_1,..., q_k)$ be two multinomial distributions over the
same index set ${1,..., k}$. If the indexes of the largest probabilities do not match on $P$ and $Q$, that is
$\argmax_i p_i \neq \argmax_j q_j$, then

\begin{align}
d_{TV}(Q,P) \geq \frac{|p_1 - p_2|}{2}
\end{align}

where $p_1$ and $p_2$ are the first and second largest probabilities in $P$.
\end{theorem}

\begin{proof}

It is easy to see that $d_{TV}(Q,P)$ is minimized when $q_1=q_2=\frac{|p_1+p_2|}{2}$ and $q_i=p_i$ for $i>=3$. This leads to the stated lower bound.

\end{proof}

Interestingly, $d_{TV}$ appears naturally in the certificates found via randomized smoothing, as a consequences of being a special case of the hockey-stick divergence. Indeed, consider a binary classification task, with a given probabilistic classifier, $f$, and an input $x$. Let $f_{c}$ denote the classifier's output at label $c$, which is the true label of $x$. Let $\mu=\mu(x)$ denote the smoothing measure on input $x$, and $\nu=\mu(x')$ denote the smoothing measure on input $x'$, with a defined distance metric $d$ such that $d(\mu, \nu)<\epsilon$. Then we can guarantee $f$ outputs the same prediction on $x'$ as on $x$ if the following is larger than $\nicefrac{1}{2}$

\begin{align}
    \min_{f_c} \E_{X\sim\nu}[f_c(X)] \quad\quad \text{ subject to }\quad\quad \E_{X\sim\mu}[f_c(X)] = p_1, 0\leq f_c(x) \leq 1
\end{align}

The dual relaxation of this problem is given by

\begin{align}
    \max_{\lambda}\lambda p_1 + \min_{0\leq f_c \leq 1}\E_{X\sim\nu}[f_c(X)] - \lambda\E_{X\sim\mu}[f_c(X)]
\end{align}

The inner minimization term is commonly referred to as the hockey-stick divergence. Since any $\lambda$ gives a valid lower bound bound to the primal problem, setting $\lambda=1$ gives

\begin{align}
    &\quad p_1 - \max_{0\leq f_c \leq 1}\E_{X\sim\mu}[f_c(X)] - \E_{X\sim\nu}[f_c(X)] \\
    &\geq p_1 - \max_{0\leq f_c \leq 1}|\int_{\mathcal{X}}f_cd(\mu-\nu) | \\
    &\geq p_1 - \max_{0\leq f_c \leq 1}\int_{\mathcal{X}}|f_c|d|\mu-\nu| \\
    &\geq p_1 - \int_{\mathcal{X}}d|\mu-\nu| \\
    &\geq p_1 - d_{TV}(\mu, \nu)
\end{align}

Thus, the classifier predicts the same label on $x'$ as on $x$ if $p_1 - d_{TV}(\mu, \nu)>\nicefrac{1}{2}$.

\section{Visualization of certified radius (for $\ell_2$ perturbations) found by $d_{\alpha}$ and $d_{\chi^2}$}
\label{app: l2_chi_renyi_compare}

\begin{figure}[t!]
\centering
  \includegraphics[width=0.75\linewidth]{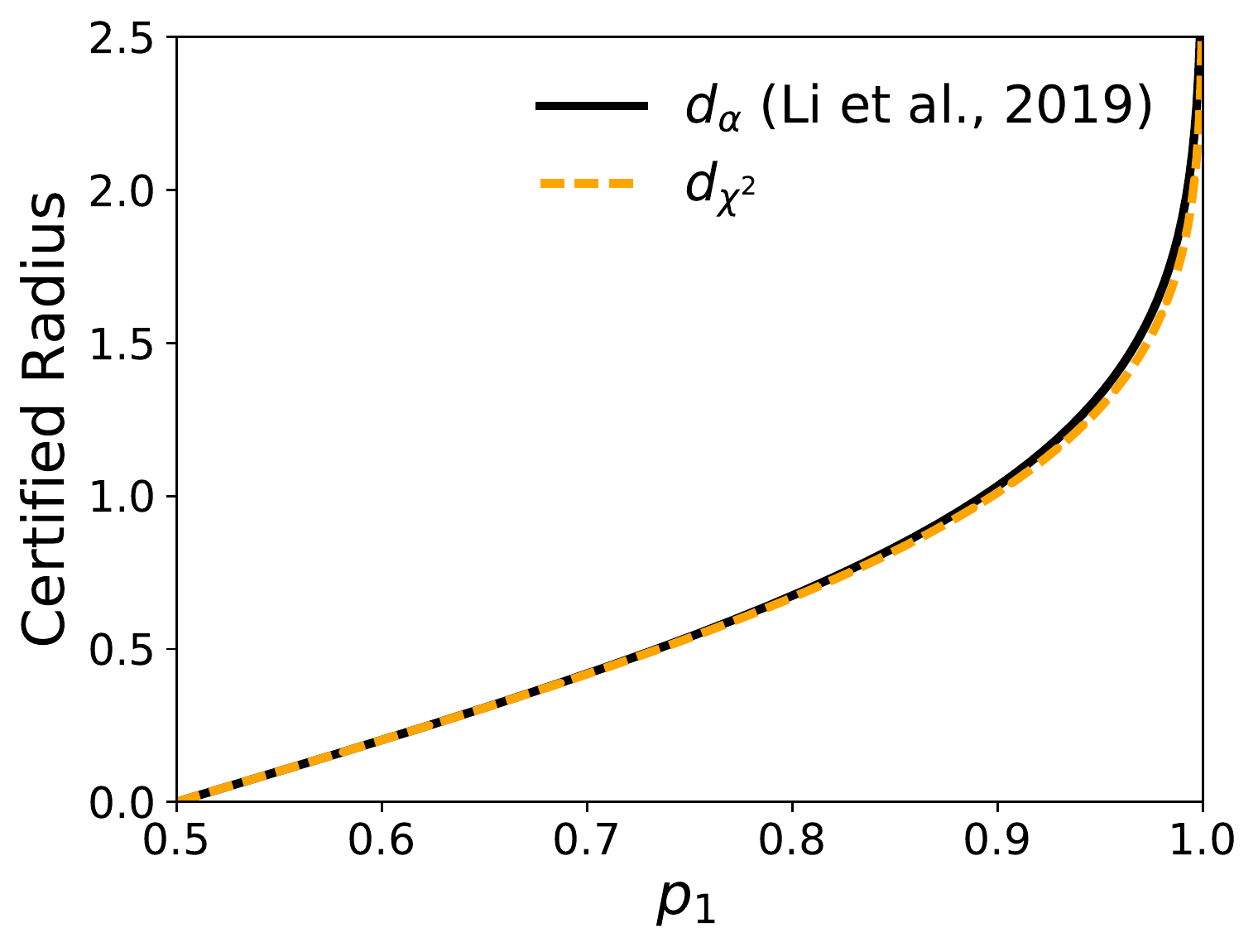}
\caption{Comparison of the certified radius against perturbations targeting the $\ell_2$ norm, for Rényi divergence ($d_{\alpha}$) and the chi-squared distance ($d_{\chi^2}$), as a function of the top predicted probability, $p_1$, with $\sigma=1$.}
\label{fig: l2_divergence_comp_renyi_chi}
\end{figure}

\Cref{fig: l2_divergence_comp_renyi_chi} visualizes the trade-off in certified radius around an input for a hypothetical binary classification task as a function of the classifier's top output probability, $p_1$. The certified radii are found using the Rényi divergence and chi-squared distance. The difference between these two certified radii is small; for $p_1\leq0.99$, the largest difference between the two radii is $0.1$.

\section{Proof of \Cref{prop_main}}
\label{app: prop_proof}

\begin{proof}

We prove this for the binary case where $p_2=1-p_1$.

\begin{enumerate}
    \item Let us fix $\alpha\in(1,\infty)$. Then $\epsilon_{d_{\alpha}} > \epsilon_{d_{\chi^2}}$ when
    \begin{align}
    & \sqrt{-\frac{2\sigma^2}{\alpha}\log\Big( 2\big(\frac{1}{2}(p_1^{1-\alpha} + (1-p_1)^{1-\alpha})\big)^{\frac{1}{1-\alpha}}\Big)} > \sqrt{\sigma^2\log(\frac{1}{4p_1(1-p_1)})} \\
    \iff& -\frac{2}{\alpha}\log\Big( 2\big(\frac{1}{2}(p_1^{1-\alpha} + (1-p_1)^{1-\alpha})\big)^{\frac{1}{1-\alpha}}\Big) > \log(\frac{1}{4p_1(1-p_1)}) \\
    \iff& 4\big(\frac{1}{2}(p_1^{1-\alpha} + (1-p_1)^{1-\alpha})\big)^{\frac{2}{1-\alpha}} < (4p_1(1-p_1))^{\alpha}
    \end{align}
    This holds $\forall p_1\in(\frac{1}{2}, 1)$ for example when $\alpha=1.1$ and so automatically holds for $\alpha \in (1,\infty)$ that maximizes the expression.
    
    \item If $\epsilon_{d_{\chi^2}} > \epsilon_{d_{KL}}$, then 
    \begin{align}
    & \sqrt{\sigma^2\log(\frac{1}{4p_1(1-p_1)})} > \sqrt{-\sigma^2\log(2\sqrt{p_1(1-p_1)})} \\
    \iff& \frac{1}{4p_1(1-p_1)} > \frac{1}{2\sqrt{p_1(1-p_1)}} \\
    \iff& (p_1 - \frac{1}{2})^2 > 0 \\
    \implies& p_1 > \frac{1}{2} 
    \end{align}
    
    \item If $\epsilon_{d_{\chi^2}} > \epsilon_{d_{H^2}}$, then 
    \begin{align}
    & \sqrt{\sigma^2\log(\frac{1}{4p_1(1-p_1)})} > \sqrt{-8\sigma^2\log(\sqrt{\frac{2-\sqrt{p_1(1-p_1)}}{2}})} \\
    \iff& \frac{1}{4p_1(1-p_1)} > \frac{2^4}{(1+2\sqrt{p_1(1-p_1)})^4} \\
    \iff& (1+2\sqrt{p_1(1-p_1)})^4 > 2^6p_1(1-p_1) \\
    \implies& p_1 > \frac{1}{2} \\
    \end{align}
    
    \item We show the inner logarithmic terms in $\epsilon_{d_{H^2}}$ and $\epsilon_{d_{B}}$ are equal, which suffices to prove equality in general. The inner logarithmic term of $\epsilon_{d_{H^2}}$ is
    \begin{align}
    & \sqrt{\frac{1+2\sqrt{p_1(1-p_1)}}{2}} \\
    =& \frac{1+2\sqrt{p_1(1-p_1)}}{\sqrt{2(1+2\sqrt{p_1(1-p_1))}}} \\
    =& \frac{(\sqrt{p_1} + \sqrt{1-p_1})^2}{\sqrt{2(1+2\sqrt{p_1(1-p_1))}}}
    \end{align}
    The last term is equal to inner logarithmic term in $\epsilon_{d_{B}}$ and so we have $\epsilon_{d_{H^2}}=\epsilon_{d_{B}}$.
    
    \item If $\epsilon_{d_{H^2}} > \epsilon_{d_{KL}}$, then
    \begin{align}
        & \sqrt{-8\sigma^2\log(\sqrt{\frac{2-\sqrt{p_1(1-p_1)}}{2}})} > \sqrt{\sigma^2\log(2\sqrt{p_1(1-p_1)})} \\
        \iff& 2^5\sqrt{p_1(1-p_1)} > (1 + 2\sqrt{p_1(1-p_1)})^4  \\
    \end{align}
    This last term has solutions in $p_1\in(\frac{1}{2}, 0.998)$.
    
    \item Let us fix $\beta \in (0, \min(1,\frac{1}{2}\log(\frac{p_1}{1-p_1}))]$, then 
    $\epsilon_{d_{KL}} > \epsilon_{\text{\cite{lecuyer2018certified}}}$ when
    \begin{align}
        & \sqrt{-\sigma^2\log(2\sqrt{p_1(1-p_1)})} > \frac{\sigma \beta}{\sqrt{2 \log (\frac{1.25(1+e^{\beta})}{p_1(1+e^{2 \beta}) - e^{2 \beta})}})} \\
        \iff& \beta^2 + 2\log (\frac{1.25(1+e^{\beta})}{p_1(1+e^{2 \beta}) - e^{2 \beta})})\log(2\sqrt{p_1(1-p_1)}) < 0 \\
    \end{align}
    This last term holds for any $p\in(\frac{1}{2}, 1)$.

\end{enumerate}

\end{proof}

\section{KL divergence of the generalized Gaussian distribution}
\label{app: gn_kl_proof}

Here, we give a proof of the claim stated in \cref{eq: gn_kl}. 

\begin{theorem}
Let $p_1$ and $p_2$ be the pdf's of two generalized Gaussians with parameters ($\mu_1$, $\sigma$, $s$) and ($\mu_2$, $\sigma$, $s$), respectively. Then $d_{KL}(p_1,p_2)$ is given by
\begin{align}
    \sum_{k=1}^{s}{s \choose k}\frac{(1+(-1)^{s-k})\Gamma(\frac{s-k+1}{s})(\mu_1 - \mu_2)^k}{2\sigma^k\Gamma(\frac{1}{s})}
\end{align}
\end{theorem}
\begin{proof}
\begin{align}
d_{KL}(p_1,p_2) &= \sum {p_1\log\Big(\frac{p_1}{p_2}\Big)} \\
&= \sum k_1e^{-|\frac{x-\mu_1}{\sigma}|^{s}}\log\Big(\frac{k_1e^{-|\frac{x-\mu_1}{\sigma}|^s}}{k_2e^{-|\frac{x-\mu_2}{\sigma}|^s}}\Big) \label{eq:kl20}
\end{align}
Where $k_1 = k_2 = \frac{s}{2\sigma\Gamma(\frac{1}{s})}$. Thus \cref{eq:kl20} is equal to 
\begin{align}
&\quad \sum k_1e^{-|\frac{x-\mu_1}{\sigma}|^{s}}\log\Big(\frac{e^{-|\frac{x-\mu_1}{\sigma}|^s}}{e^{-|\frac{x-\mu_2}{\sigma}|^s}}\Big)  \\
&= \mathbb{E}_{p_1}\bigg[\Big(\frac{x-\mu_2}{\sigma}\Big)^s - \Big(\frac{x-\mu_1}{\sigma}\Big)^s\bigg]  \\
&= \frac{1}{\sigma^s}\mathbb{E}_{p_1}\big[(x-\mu_2)^s - (x-\mu_1)^s\big] \label{eq:kl21}
\end{align}

Note that $(x-\mu_2)^s = \sum_{k=0}^{s}{s \choose k}x^{s-k}(-\mu_2)^k$. Thus \cref{eq:kl21} is equal to

\begin{align}
\begin{split}
    &\frac{1}{\sigma^s}\bigg[\Big(\sum_{k=0}^{s}{s \choose k}{\mu_1}^{s-k}(-\mu_2)^k\sum_{i=0}^{s-k}{s-k \choose i}(\frac{\sigma}{\mu_1})^{i}(1+(-1)^i)\frac{\Gamma(\frac{i+1}{s})}{2\Gamma(\frac{1}{s})}\Big) \\
    &\quad -\Big(\sum_{k=0}^{s}{s \choose k}{\mu_1}^{s-k}(-\mu_1)^k\sum_{i=0}^{s-k}{s-k \choose i}(\frac{\sigma}{\mu_1})^{i}(1+(-1)^i)\frac{\Gamma(\frac{i+1}{s})}{2\Gamma(\frac{1}{s})}\Big)\bigg]  \\
    \end{split} \\
\begin{split}
    &= \frac{1}{\sigma^s}\bigg[(\sum_{k=0}^{s}{s \choose k}{\mu_1}^{s-k}(-\mu_2)^k  \\
    &\quad- \sum_{k=0}^{s}{s \choose k}{\mu_1}^{s-k}(-\mu_2)^k\sum_{i=1}^{s-k}{s-k \choose i}(\frac{\sigma}{\mu_1})^{i}(1+(-1)^i)\frac{\Gamma(\frac{i+1}{s})}{2\Gamma(\frac{1}{s})}  \\
    &\quad -\sum_{k=0}^{s}{s \choose k}{\mu_1}^{s-k}(-\mu_1)^k  \\
    &\quad- \sum_{k=0}^{s}{s \choose k}{\mu_1}^{s-k}(-\mu_1)^k\sum_{i=1}^{s-k}{s-k \choose i}(\frac{\sigma}{\mu_1})^{i}(1+(-1)^i)\frac{\Gamma(\frac{i+1}{s})}{2\Gamma(\frac{1}{s})}\bigg]  \\
\end{split} \\
\begin{split}
    &= \frac{1}{\sigma^s}\bigg[(\mu_1-\mu_2)^s  \\
    &\quad+ \sum_{k=0}^{s}{s \choose k}{\mu_1}^{s-k}(-\mu_2)^k\sum_{i=1}^{s-k}{s-k \choose i}(\frac{\sigma}{\mu_1})^{i}(1+(-1)^i)\frac{\Gamma(\frac{i+1}{s})}{2\Gamma(\frac{1}{s})} \\
    &\quad- \sum_{k=0}^{s}{s \choose k}{\mu_1}^{s-k}(-\mu_1)^k\sum_{i=1}^{s-k}{s-k \choose i}(\frac{\sigma}{\mu_1})^{i}(1+(-1)^i)\frac{\Gamma(\frac{i+1}{s})}{2\Gamma(\frac{1}{s})}\bigg] \label{eq: kl28}
\end{split}
\end{align}

Note that only even indices contribute to the summand in \cref{eq: kl28} because of the
$(1+(-1)^i)$ term and so can be written as

\begin{align}
& \frac{1}{\sigma^s}(\mu_1-\mu_2)^s 
+ \frac{1}{\sigma^s}\Big(\sum_{k=1}^{s}{s \choose k}({\mu_1}^{s-k}(-\mu_2)^k-{\mu_1}^{s-k}(-\mu_1)^k)\sum_{i>0}^{s-k}{s-k \choose i}(\frac{\sigma}{\mu_1})^{i}(1+(-1)^i)\frac{\Gamma(\frac{i+1}{s})}{2\Gamma(\frac{1}{s})}\Big) \label{eq: kl4}
\end{align}

Note, $k=0 \implies ({\mu_1}^{s-k}(-\mu_2)^k-{\mu_1}^{s-k}(-\mu_1)^k)=0$, and so \cref{eq: kl4} becomes

\begin{align}
\sum_{k=1}^{s}{s \choose k}\frac{(1+(-1)^{s-k})\Gamma(\frac{s-k+1}{s})(\mu_1 - \mu_2)^k}{2\sigma^k\Gamma(\frac{1}{s})}
\end{align}

\end{proof}

\newpage

\section{How does $\sigma$ affect the certification radius?}
\label{app: sigma_certify}

\begin{figure}[t]
\centering
  \includegraphics[width=0.75\columnwidth]{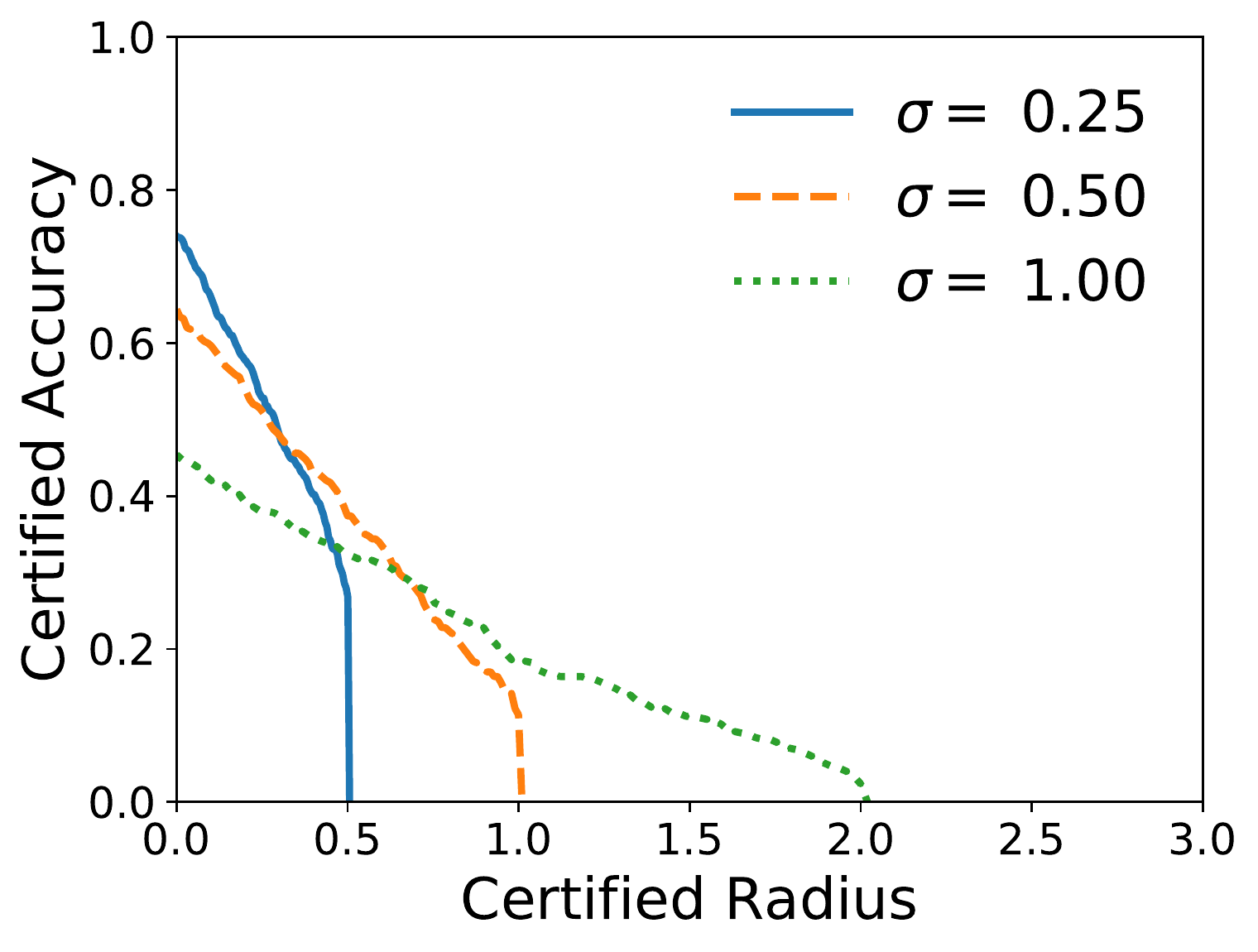}
  \label{fig:sub2}
\caption{Certified accuracy against perturbations targeting the $\ell_2$ norm for CIFAR-10. Given as a function of the certified radius, the radius around which an input is robust.}
\label{fig:sigma_cert}
\end{figure}

For 400 CIFAR-10 test set inputs, we certify inputs against $\ell_2$ perturbations while varying the
noise scale parameter $\sigma$~\footnote{Note, sampling from a generalized Gaussian distribution with scale $\sigma$ and shape $s=2$, is equivalent to sampling from a Gaussian distribution with scale $\nicefrac{\sigma}{\sqrt{2}}$.}.
~\Cref{fig:sigma_cert} shows 
certified accuracy as a function
of the certified area
for $\sigma=0.25, 0.5, 1.0$. This is the guaranteed classification accuracy under any perturbation smaller than the specified bound. Larger $\sigma$ results in a larger 
certified area but suffers from lower standard classification accuracy -- this corresponds to accuracy under a certified radius of 0. This mirrors the findings of Cohen \etal \cite{cohen2019certified} and Tsipras \etal \cite{tsipras2018there} who showed a trade-off between robustness and standard accuracy.

\section{Samples smoothed with different forms of generalized Gaussian noise}
\label{app: viz_samples}

\begin{figure}[t]
\centering
        \begin{subfigure}[t]{0.24\textwidth}
                \includegraphics[width=0.95\linewidth]{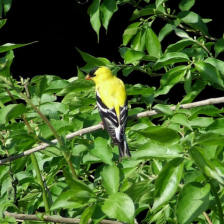}
                \caption{Goldfinch}
                \label{fig:standard_cifar10}
        \end{subfigure}%
        \begin{subfigure}[t]{0.24\textwidth}
                \includegraphics[width=0.95\linewidth]{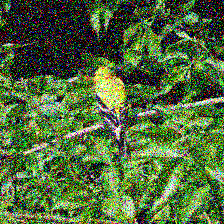}
                \caption{$\epsilon = 0.98$ ($\ell_1$)}
                \label{fig:l1_cifar10}
        \end{subfigure}%
        \begin{subfigure}[t]{0.24\textwidth}
                \includegraphics[width=0.95\linewidth]{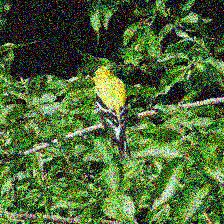}
                \caption{$\epsilon = 0.44$ ($\ell_2$)}
                \label{fig:l2_cifar10}
        \end{subfigure}%
        \begin{subfigure}[t]{0.24\textwidth}
                \includegraphics[width=0.95\linewidth]{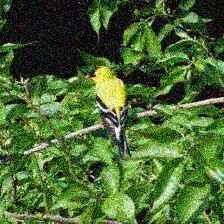}
                \caption{$\epsilon = 0.29$ ($\ell_3$)}
                \label{fig:l3_cifar10}
        \end{subfigure}
        \\
        
        \begin{subfigure}[b]{0.24\textwidth}
                \includegraphics[width=0.95\linewidth]{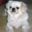}
                \caption{Dog}
                \label{fig:standard_imagenet}
        \end{subfigure}%
        \begin{subfigure}[b]{0.24\textwidth}
                \includegraphics[width=0.95\linewidth]{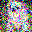}
                \caption{$\epsilon = 0.95$ ($\ell_1$)}
                \label{fig:l1_imagenet}
        \end{subfigure}%
        \begin{subfigure}[b]{0.24\textwidth}
                \includegraphics[width=0.95\linewidth]{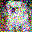}
                \caption{$\epsilon = 0.48$ ($\ell_2$)}
                \label{fig:l2_imagenet}
        \end{subfigure}%
        \begin{subfigure}[b]{0.24\textwidth}
                \includegraphics[width=0.95\linewidth]{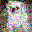}
                \caption{$\epsilon = 0.32$ ($\ell_3$)}
                \label{fig:l3_imagenet}
        \end{subfigure}
        \caption{Two randomly chosen images from ImageNet (Top) and CIFAR-10 (Bottom). We give examples of noise from a generalized Gaussian distribution with $s=1, 2, \text{ and } 3$, and the maximum perturbation size, $\epsilon$, for which the classifier is certified to predict the correct class under $\ell_1, \ell_2, \text{ and } \ell_3$ based attacks. }\label{fig:cert_example}
\end{figure}

In \cref{fig:cert_example}, we visualize the smoothing of a generalized Gaussian over two random inputs from CIFAR-10 and ImageNet test sets. \Cref{fig:standard_cifar10,fig:standard_imagenet} correspond to the non-smoothed versions of these two inputs, \cref{fig:l1_cifar10,fig:l1_imagenet} correspond to the inputs smoothed with generalized Gaussian noise sampled from $\mathcal{GN}(0, 0.25, 1)$. Similarly, \cref{fig:l2_cifar10,fig:l2_imagenet} correspond to the inputs smoothed with generalized Gaussian noise sampled from $\mathcal{GN}(0, 0.25, 2)$, and \cref{fig:l3_cifar10,fig:l3_imagenet} correspond to the inputs smoothed with generalized Gaussian noise sampled from $\mathcal{GN}(0, 0.25, 3)$. For each smoothed input, we state the size of the certified radius, $\epsilon$ -- upto this value the input is robust to adversarial perturbations in the specified $\ell_p$ norm.

\section{An example of separability of optimal decision boundaries for different $\ell_p$ norms}
\label{sec: lp_sphere_example}

Khoury \etal \cite{khoury2018geometry} hypothesize that, in general, it is impossible for a classifier to be robust against all $\ell_p$ norm attacks. They consider a toy example to demonstrate this: consider two $n$-dimensional spheres, $\mathrm{X}_1$ and $\mathrm{X}_2$, both centered at the origin with radii $r_1$ and $r_2$, respectively. They show that the optimal decision boundary between points on the spheres are distinct under the $\ell_2$ and $\ell_\infty$ norms. We extend this to arbitrary norms through \Cref{thm: lp_sphere_example_theorem}. First, we  define what we mean by an optimal decision boundary, state the conjecture and then give a draft of a proof that decision boundary separability extends to other norms. 

\begin{definition}
 Let $\Delta$ be a set of points in $\mathbb{R}^n$. We say $\Delta$ \emph{separates}
$\mathrm{X}_1$ and $\mathrm{X}_2$ if any continuous function $f$ that passes through $\mathrm{X}_1$ and $\mathrm{X}_2$ also passes
through $\Delta$.
\end{definition} 

\begin{definition}
  Let $\Delta$ be any separator of $\mathrm{X}_1$ and $\mathrm{X}_2$. Choose
a point $x\in\Delta$ and consider the ball $\mathcal{B}_{\epsilon, p}(x) := \{ {z} \mid  {\epsilon \geq \norm{z-x}_p } \}$. We call $\Delta$
a maximum separator if $\forall x \in \Delta$ the following holds: $\forall \epsilon > 0, \exists m_1, m_2 \in \mathcal{B}_{\epsilon, p}(x)$,
and points $y_1,y_2 \in \mathrm{X}_1\bigcup\mathrm{X}_2$, such that if $y_1$ is the point that minimizes
$\norm{m_1 - y}_p$ (where $y\in \mathrm{X}_1\bigcup\mathrm{X}_2$), then $y_1 \in \mathrm{X}_1$, and equivalently if 
$y_2$ is the point that minimizes $\norm{m_2 - y}_p$ then $y_2 \in \mathrm{X}_2$.
\end{definition}

\noindent \textbf{An example of a separator that is not maximal.} Let $\Delta = \mathrm{X}_1$. Then $\exists x \in \mathrm{X}_1$ and 
$\epsilon > 0$ such that $\forall z \in \mathcal{B}_{\epsilon}(x)$, the points $y\in \mathrm{X}_1\bigcup\mathrm{X}_2$ that minimize
$\norm{z-y}_p$ all lie on $\mathrm{X}_1$ (i.e. $y\in \mathrm{X}_1$ and $y \notin \mathrm{X}_2$).

\begin{conjecture}
\label{thm: lp_sphere_example_theorem}
Let two concentric spheres $\mathrm{X}_1, \mathrm{X}_2 \in \mathbb{R}^n$ have radii $r_1, r_2$, respectively.
Then $\forall p, q \geq 1$ with $p \neq q$, 
$\Delta_p \neq \Delta_q$, where $\Delta_p$ denotes the maximal separator in the $\ell_p$ norm. 
\end{conjecture}

We give a `proof by example' in two dimensions, showing that $\Delta_1 \neq \Delta_2 \neq \Delta_4 \neq \Delta_\infty$, and prove that $\Delta_1 \neq \Delta_2 \neq \Delta_\infty$ in $n$-dimensions. First, consider concentric circles $\mathrm{X}_1, \mathrm{X}_2 \in \mathbb{R}^2$ with radii $1, 4$, respectively.

$\Delta_2$ defines a circle of radius $\frac{5}{2}$. In particular for $p=(x,y)$, when $x=0$, $p\in \Delta_2$ has $y$-coordinate
$\frac{5}{2}$. For $\Delta_\infty$, when $x=0$, $p$ has $y$-coordinate $\frac{3+\sqrt{79}}{5}$. To see this, $\mathcal{B}_{\epsilon, \infty}(m)$
with center $m=(0, 1+\kappa)$ touches $\mathrm{X}_2$ at $q=(\kappa, 1+2\kappa)$. At $q$ we have $\kappa^2 + (1+2\kappa)^2 = 4^2$, and so
$\kappa = \frac{-2 + \sqrt{79}}{5}$. Hence, at $x=0$, $q \in \Delta_\infty$ has $y$-coordinate $\frac{3+\sqrt{79}}{5}$.

To find $y$-coordinate when $x=0$ for a point $q \in \Delta_4$, we must solve 

\begin{align}
    & x^2 + y^2 = 4^2 \label{eqn:sphere_l4_example_1}\\ 
    & x^4 + (y-(1+\kappa))^4=\kappa^4 \label{eqn:sphere_l4_example_2} 
\end{align}

Since $\Delta_4$ is tangential to $\mathrm{X}_2$, we must find the root of the determinant of $(4^2 - y^2)^2 + (y-(1+\kappa))^4 - \kappa^4 = 0$.
This an order 12 polynomial, 

\begin{align}
\begin{split}
    & 28\kappa^{12} + 96\kappa^{11} + 176\kappa^9 - 4540\kappa^8 - 19528\kappa^7 +\\
    & 15916\kappa^6 + 
    403800\kappa^5 + 495735\kappa^4 - 3757020\kappa^3 + \\
    & 3592350\kappa^2 + 16024500\kappa - 24350625 = 0. \label{eqn:sphere_l4_example_3} 
\end{split}
\end{align}

This has no solution in the radicals and is approximately $1.4755$ and so $q \in \Delta_4$ has $y$-coordinate $2.4755$.

To find $\Delta_p$ in general we must solve high order polynomials that may not factor. However, we can find $\Delta_1$ in $n$ dimensions. Consider the diamond $\ell_1$ ball centered at 
$m = (\frac{r_1}{2} + \frac{\kappa}{2}, \frac{r_1}{2} + \frac{\kappa}{2},...,\frac{r_1}{2} + \frac{\kappa}{2})$, and 
$q\in \Delta_1$ has coordinate $(\frac{r_1}{2} + \frac{\kappa}{2} + \kappa, \frac{r_1}{2} + \frac{\kappa}{2},..., \frac{r_1}{2} + \frac{\kappa}{2})$.
Then $(n-1)(\frac{r_1}{2} + \frac{\kappa}{2})^2 + (\frac{r_1}{2} + \frac{\kappa}{2} + \kappa)^2 = r_2^2$. Thus,

\begin{align}
    & \kappa = -\frac{n+2}{n+8}\sqrt{2}r_1 + \frac{2}{n+8}\sqrt{(n+8)r_2^2 - 2(n-1)r_1^2}. \label{eqn:sphere_l1_example}  
\end{align}

Thus, similarly to Khoury \etal \cite{khoury2018geometry}, for constant $r_1$ and $r_2$, $\Delta_1$ scales like $\mathcal{O}(\frac{1}{\sqrt{n}})$, and for a classifier
trained to learn $\Delta_1$, an adversary can construct an adversarial perturbation in the $\ell_2$ norm as small as $\mathcal{O}(\frac{1}{\sqrt{n}})$.

\end{appendix}

\end{document}